\documentclass[authoryear,1p]{elsarticle}
\usepackage[colorlinks=true,linkcolor=black, citecolor=blue, urlcolor=blue]{hyperref}
\usepackage{amsmath}

\usepackage[inline]{enumitem}
\newlist{inline}{enumerate*}{1}
\setlist[inline]{label=(\roman*)}

\usepackage{amssymb}

\usepackage{float}
\usepackage{multirow}

\usepackage{listings}
\journal{Transportation Research Part C: Emerging Technologies}

\begin{document}

\begin{frontmatter}



\title{Black box behavioural modelling: Predicting human activity schedules with a deep conditional generative approach}


\author[inst1]{Fred Shone}\ead{fredjshone@gmail.com}
\author[inst1]{Tim Hillel\corref{cor1}}\ead{tim.hillel@ucl.ac.uk}
\cortext[cor1]{Corresponding author}
\affiliation[inst1]{organization={Department of Civil, Environmental and Geomatic Engineering, University College London},
    country={UK}}


\begin{abstract}

Modelling the complexity and diversity of human activity scheduling behaviour is inherently challenging. We demonstrate ActVAE, a deep conditional-generative machine learning approach for the modelling of activity schedules. Suitable for application in activity-based demand modelling frameworks, schedules are modelled as conditional on individual, household and schedule information, such as age, income, and access to public transit. We demonstrate the rapid generation of precise, realistic and diverse schedules dependent on input labels. We extensively evaluate and compare model capabilities against baseline models using a joint-density estimation framework. In addition to providing a novel alternative to existing scheduling approaches, our work highlights the value of explicitly modelling the randomness of complex and diverse human behaviours.
\end{abstract}


    


\end{frontmatter}




\section{Introduction}
\label{sec:intro}

We consider the modelling of human activity schedules based on input labels, such as household size, car ownership and access to public transit. For this work, we define an activity schedule as a sequence of some number of activity participations, with associated start times and durations.

Human activity schedules result from a complex combination of processes, including:
\begin{inline}
    \item constraints, such as needing to fit all activities within a given time budget,
    \item preferences, such as wanting to visit a friend or spend longer shopping, and
    \item interactions, both with each other and the environment, such as within a household or with traffic.
\end{inline}
Modelling realistic and representative activity schedules is therefore inherently challenging.

Activity scheduling is a core requirement within activity-based demand modelling approaches. Activity scheduling can be considered as the prediction, for an individual, of activity participations and either their timings, or their order and durations within a given time-frame. The dominant approaches in practice, originally by \cite{bowman1998}, decompose activity scheduling into compositions of sub-models. These complex compositions are slow and expensive to develop, likely resulting in - as per \cite{Miller} - “the (very) slow adoption of activity-based travel demand models in operational planning practice”.

Our primary contribution is to introduce an activity scheduling model called ActVAE. We intend ActVAE to be suitable for application as the scheduling component within a broader activity-based demand modelling framework, for predicting activity participations and when they occur. In addition to the practical convenience of combining multiple components into a single fast scheduling model, we show that our approach allows for more diverse and realistic scheduling compared to existing approaches, especially for the timing of activities. ActVAE schedules also have high temporal fidelity and are temporally consistent.

ActVAE is systematically evaluated against a series of three benchmarks, including a conventional \emph{compositional} modelling approach. We show that ActVAE is able to best generate realistic distributions of schedules conditional on explanatory variables.

ActVAE is a structured latent generative approach, based around a Variational Auto-Encoder (VAE) design. Compared to previous schedule \emph{synthesis} work using a VAE \citep{SELF}, we add conditional capability, aligning the approach with existing \emph{modelling} approaches and allowing direct comparison.

We additionally quantify the importance of conditionality relative to unexplained variation for modelling schedules. Perhaps unsurprisingly, given the complexity of our environment, interactions, and human decision-making processes, we show that unexplained variation dominates. This finding strongly supports the application of generative approaches that explicitly capture this variation, both for activity scheduling and for other complex human behaviours more generally.

All experiments are reproducible using code added to the open-source project Caveat\footnote{https://github.com/big-ucl/caveat}.

\section{Literature review}
\label{sec:lit}

We review existing activity scheduling approaches, summarised in Table \ref{tab:models}. Note that many of the compared approaches are full activity-based demand modelling frameworks, so we restrict comparison to the components of these models responsible for activity scheduling only. We further present a short primer on deep generative learning, summarise existing applications of generative learning for activity scheduling, and finally give a summary of the recent developments in generative learning relevant to this research.

\begin{table}
    \footnotesize
    \caption{Taxonomy of existing activity scheduling approaches}
    \vspace{2ex}
    \centering
        \begin{tabular}{l | l | l l l}
        \hline
        Model & Approach & \multicolumn{3}{c}{Taxonomy} \\
        \hline
        ALBATROSS* \citep{ALBATROSS} & heuristic & theory-driven & compositional & calibrated generative\\
        CEMPDAP* \citep{CEMDAP} & econometric & theory-driven & compositional & calibrated generative\\
        TASHA* \citep{TASHA} & mixed & theory-driven & compositional & calibrated generative\\
        DaySim* \citep{DaySim} & econometric & theory-driven & compositional & calibrated generative\\
        CT-RAMP* \citep{ct-ramp} & econometric & theory-driven & compositional & calibrated generative\\
        ActiTopp* \citep{actiTopp} & econometric & theory-driven & compositional & calibrated generative\\
        WGAN \citep{badumarfo2020differentially} & ML & data-driven & joint & implicit generative \\
        SALT* \citep{trees} & ML & data-driven & compositional & calibrated generative\\
        TripGAN \citep{tripCGAN} & ML & data-driven & joint & implicit generative \\
        SDS* \citep{SDS} & mixed & theory-driven & compositional & calibrated generative\\
        OASIS \citep{POUGALA2023104291} & optimisation & theory-driven & joint & calibrated generative\\
        ActivitySim* \citep{activitysim} & econometric & theory-driven & compositional & calibrated generative\\
        Conditional RNN \citep{koushikActivityScheduleModeling2023} & ML & data-driven & joint & explicit generative \\
        Act2Loc \citep{act2loc} & ML & data-driven & joint & explicit generative \\
        ActVAE (this paper) & ML & data-driven & joint & explicit generative \\
        \hline
        \multicolumn{5}{l}{}\\[-1em]
        \multicolumn{5}{l}{{\small \textasteriskcentered{ Activity-based demand model, we consider activity scheduling components only}}} \\
        
    \end{tabular}           
    \label{tab:models}
\end{table}

\subsection{Traditional compositional approaches}
\label{sec:compositional_approaches}

The current prevalent approach to activity scheduling, originally proposed by \cite{bowman1998}, decomposes scheduling into a series of choice components. The exact structure and nature of decompositions vary. The literature typically distinguishes between 
\begin{inline}
    \item econometric-based frameworks, such as CEMDAP \citep{CEMDAP}, DaySim \citep{DaySim}, CT-RAMP \cite{ct-ramp}, ActiTopp \citep{actiTopp}, and ActivitySim \citep{activitysim},
    \item heuristic or rules-based approaches, such as ALBATROSS \citep{ALBATROSS}, and
    \item mixtures of the two, such as TASHA \citep{TASHA} and SDS \citep{SDS}.
\end{inline}

The structure of decompositions is a heuristic decision, influenced by theory, practicality and convention, so we classify these models as \emph{theory-driven} and \emph{compositional}. Compositional models are specified, estimated, evaluated and calibrated component-by-component, with later components conditional on the outputs of those prior. Combined, this creates the capacity to generate novel schedules and can support the learning of reasonably complex joint distributions. However, the final output distribution requires manual calibration of components and structure, and so we classify these models as \emph{calibrated generative}.

In isolation, model components, particularly discrete choices, such as primary activity type, are highly interpretable. But in combination, these systems are complex. To control complexity decompositions simplify real choices, for example limiting participation sequences to common tour types, and discretising time steps into periods. This limits the capacity to represent the real distribution of activity scheduling.

\subsection{Joint approaches}

Some research seeks to reduce decompositions through partially or fully joint scheduling models. Specifically, we consider approaches as \emph{joint} if they can be estimated and calibrated end-to-end, for example, using a single loss function.

Joint approaches to activity scheduling reduce complexity and potentially allow for better representation of joint distributions within schedules. \cite{BHAT2005, BHAT2008} demonstrate a multiple discrete-continuous extreme value (MDCEV) approach that simultaneously models activity participations and durations. However, additional sequencing or timing components are still required to form complete activity schedules.

The Household Activity Pattern Problem (HAPP) \citep{Recker1995} formulates household activity scheduling as an optimisation problem, where households seek to achieve prescribed activities while minimising travel disutility, without considering activity utility. \cite{ChowRecker2012} extend HAPP with an inverse optimisation procedure to calibrate its parameters against observed travel diary data.

Building on HAPP, OASIS \citep{POUGALA2023104291} jointly models activity schedules including participations (and locations and modes) consistent with behavioural theory using a utility optimisation approach. However, estimation of parameters and simulation of schedules is computationally expensive, limiting scalability. \cite{Manser2021ResolvingTS} manage to scale the approach to application as part of an activity-based transport demand model, but they limit the scope of the simultaneous approach to activity timings only.

\subsection{Machine learning approaches}

Initial machine learning (ML) based approaches to activity scheduling have also adopted \emph{compositional} approaches, replacing the theory-driven components with more \emph{data-driven} alternatives, such as classification models, e.g. for sequence prediction \citep{dap_ml} and duration prediction \citep{trees}.

ML approaches to discrete choice problems have been shown to scale to massive choice sets \citep{rumboost}. This makes the combination of decomposed choices, such as location and mode, into single joint or simultaneous choices feasible. However, this has not yet been extended to activity schedule modelling, which can be thought of as the simultaneous choice of all daily activity participations and their timings.

\subsection{ A primer to deep generative modelling}

Deep Generative Models (DGMs) make use of large training datasets and deep machine learning architectures to learn the distributions of complex high-dimensional data such as images \citep{attngan}, video \citep{nova}, and text \citep{bert}.

Generative models can be either conditional-generative or purely generative. Conditional-generative models use explanatory variables to influence or control the generation process. Purely generative approaches do not utilise explanatory variables, and hence their generative processes cannot be controlled. This is sometimes referred to as \emph{synthesis} rather than \emph{modelling}.

DGMs often require huge data and compute budgets to train. \emph{Data efficiency} is the ability of a model to learn from relatively little data. Activity schedule data is of limited scale compared to some other domains, so data efficiency is important. Similarly, \emph{computational efficiency} is the ability of a model to be trained or to generate relatively quickly or with low computational cost.

Probability \emph{density estimation} is a measure of how well the generated distribution of a model matches the desired distribution. Good density estimation is a general requirement of all generative models. Density estimation can be decomposed into intra- and inter-sample density estimation. Intra-sample density considers the relationships between features within individual samples, for example, images of people are expected to have the correct number of fingers, and text is expected to be grammatically correct. In the context of activity scheduling, intra-sample distributions influence the structural feasibility of schedules and also the relationships between choices made across schedules. Inter-schedule distributions consider the aggregate patterns of schedules.

In many applications, intra-sample distribution, is prioritised. This is often described as a focus on \emph{sample quality} rather than overall density estimation. In contrast, both intra and inter-sample density estimation are important in our context, as we wish to use schedules to represent the choices of real populations both in aggregate, rather than just individually.

\subsection{Existing deep generative modelling approaches to activity scheduling}

Existing research into DGM approaches to activity scheduling have predominantly been with Generative Adversarial Networks (GANs) and, most recently, with autoregressive approaches. GANs use a discriminator model to guide the training of a generator model, potentially allowing for the learning of very complex distributions conditional on input labels. However, this \emph{implicit} approach to modelling the distribution of data is prone to poor estimation of the inter-sample variance of outputs. This limitation is likely why work using GANs by both \cite{badumarfo2020differentially} and \cite{tripCGAN} acknowledge the poor quality of more complex modelled distributions.

Perhaps inspired by progress in the language modelling domain, \cite{koushikActivityScheduleModeling2023} and \cite{act2loc} model activity schedules using auto-regressive approaches. However, autoregressive approaches, particularly for long sequences, are data and compute inefficient, which, despite their great success in domains such as text generation and LLMs, makes them unsuitable for activity-schedule generation. This likely results in poor density estimation noted by \cite{koushikActivityScheduleModeling2023}, and temporal aggregations and other simplifications by \cite{act2loc} which limit the realistic diversity of generated schedules.

The above works are conditional-generative. Schedules are conditioned on socio-economic attributes to potentially enable the realistic joint distribution of schedules with socio-economic labels. However, there is very limited evaluation of the quality of either the generative or conditional processes, so there is little assurance that the models will perform well in practice.

\cite{SELF} demonstrate a Variational Auto-Encoder (VAE) \citep{vae} for data-efficient and explicit consideration of both the intra and inter-sample distribution of schedules. Further to the above work, they propose a density estimation framework to robustly demonstrate that the distribution of generated schedules is realistic. However, their approach is purely generative, thus the model cannot be used to generate activity schedules dependent on input attributes. Conditionality is a crucial element of transport demand modelling, used to model responses to new scenarios such as changes in demographics or transport accessibility. The relationship between individuals and their schedules is also important for considering the equity of scenarios.

\subsection{Relevant developments in deep generative modelling}

Progress in DGMs is dominated by Diffusion \citep{diffusion} for images and autoregression for language. However, much of this work presumes the availability of huge volumes of training data. In more data-constrained domains, such as tabular data \citep{tabularVAE}, 3D imaging \citep{3dimage}, and medicine \citep{VAEaugment}, VAEs are popular because, as per \cite{data_efficiency}, they are more data-efficient than other approaches.

The VAE architecture also allows for tractable estimation of likelihood. During training, this results in the model explicitly assigning likelihood to all training data. As discussed by \cite{GANdensity} and demonstrated by \cite{precandrecal}, this results in better density estimation than alternative approaches, particularly GANs. VAEs are therefore popular for data augmenting tasks \citep{EndoVAE, HVAE, vetVAE} where maintaining the original distribution in a data-efficient manner is critical.

The original VAE architecture is generative-only. Various approaches exist for incorporating conditionality to create Conditional VAE (CVAE) architectures. The base approach, originally described by \cite{cvae}, is to \emph{inject} conditioning information into the VAE architecture using addition, concatenation, or other operations such as FILM by \cite{film}.

The above CVAE approaches all rely on the model to learn to disentangle useful conditional information from random variance \citep{betaVAE}. In practice, especially where conditional information is weak or noisy, CVAEs suffer from conditional collapse, where the model learns to ignore conditional information, making generation uncontrollable.

Conditional labels for activity schedules, such as age or income, are particularly weak, providing some, but not much information about scheduling behaviour. This uncertainty is problematic. DGM literature treats such uncertainty as noise or mislabelling, but in the case of scheduling and human behavioural modelling more generally, this noise is aleatoric uncertainty inherent to the real schedule generation process.

A variety of architectures have been proposed to improve CVAE disentangling and tackle conditional collapse. These fall into two types; \begin{inline}
    \item auxiliary classification, and
    \item conditional priors
\end{inline}. The former use the loss from an auxiliary classification model to encourage the CVAE to generate controllable outputs. This auxiliary classifier can be used on the CVAE latent as per \cite{kingma_cvae} or decoder output as per \cite{acvae}. A conditional prior, such as CP-VAE by \cite{cpvae2}, instead seeks to make the CVAE prior, which is otherwise a standard isotropic Gaussian, more expressive by making it a function of the conditional information using a prior network. CP-VAE is discussed in detail in \ref{sec:cpvae}.

\section{Motivations for our approach}
\label{sec:motivations}

\subsection{Informal problem definition}

We seek to predict an individual's activity schedule, i.e. their sequence of activity participations (such as home or work) and when they will occur. Predictions are modelled dependent on input labels. Labels may belong to the individual undertaking the activity schedule, such as age, their household, such as income, or attributes more broadly associated with the activity schedule, such as weather conditions or transport connectivity. 

Predicted schedules should be correctly structured, in particular temporally consistent, such that schedules do not have overlapping activity participations or gaps. Predicted schedules should also be realistically controllable by explanatory labels so that new scenarios can be modelled, for example, by increasing the age of individuals or improving their connectivity. Schedules are observed to be highly varied, and only some of this variation can be attributed to the explanatory labels. The distribution of predicted schedules should also realistically reflect this uncertainty.

The treatment of activity scheduling as a generative problem is foundational to our approach to modelling schedule variance. In the following, we detail why and how activity scheduling should be considered as a generative problem. We then provide theoretical support for the good performance of our chosen architecture. Key notation for this section is summarised in Table \ref{tab:gen_gloss}.

\subsection{Generative versus discriminative problems}
\label{sec:challenge}

\begin{table}
    \footnotesize
    \caption{Glossary of probability notation for Section \ref{sec:motivations}}
    \vspace{2ex}
    \centering
        \begin{tabular}{l l }
        \hline
        $x$ & schedule \\
        $y$ & labels (conditional data for schedule) \\
        $p(x)$ & probability distribution of schedules \\
        $p(x, y)$ & joint distribution of schedules and labels \\
        $p(x \mid y)$ & conditional distribution of schedules on labels \\
        $z$ & a latent embedding or representation \\
        $p(z)$ & the latent prior distribution \\
        \hline
    \end{tabular}           
    \label{tab:gen_gloss}
\end{table}

Discriminative models, such as those used to represent individual choice components in compositional activity scheduling frameworks, seek to directly model the conditional probability $p(y | x)$, where $y$ is an output label/s and $x$ some input feature/s. For example, we might seek to predict if an input schedule belongs to someone who has access to a car, where $x$ is a schedule, and $y$ is a car ownership label. However, for this work, we wish to model the reverse, $p(x | y)$, i.e. the probability of a schedule dependent on input labels, such as car ownership. The key distinction between labels~$y$ and schedules~$x$ is that all possible values of the labels are known and defined, whereas that is not the case for all possible schedules. In the first case, where we seek to predict an output label, available data typically enumerates all possible output labels, but in the latter case, where we seek to predict schedules, available data does not enumerate all possible schedules. We use this \textit{reversed} notation throughout, i.e. $p(x | y)$, consistent with generative machine learning literature such as \cite{vae}.

Consider a schedule aggregation of one-hour time steps and only two possible activity types. There are then $2^{24}$ (over 16 million) possible 24-hour activity schedules. More precise schedule representations, incorporating more activity types and finer temporal steps, quickly approach an infinite choice-set size. In comparison, available datasets of human activity schedules typically contain only tens of thousands of samples. An important consideration for any modeller is therefore whether they wish to generate schedules not previously observed in the limited training data. To do so requires learning the distribution $p(x \mid y)$, i.e. the probability distribution of all possible schedules. This is a distinctly \emph{generative modelling} problem.

\subsection{Critique of compositional approaches}
\label{sec:parallels}
 
 Compositional models are estimated component by component and then manually calibrated. The resulting distribution of outputs is heavily dependent on the ability of the modeller to manually specify and estimate each component and choose their combined structure. In practice, compositional models are calibrated to target key metrics, such as trip rates. More precise distributions, particularly those created by multiple interacting decompositions, are either not measured or, if they are, are challenging to address. This makes model and scenario development slow, expensive, and specialised.

Additionally, decompositions are limited and simplified, limiting the realism of of output schedules and their distributions. For example, sequences are commonly restricted to common patterns and time aggregated into course time steps. In some applications, simplification is not problematic, for example, if modelling aggregate travel demand for morning and evening peaks. However, increasingly, realism, particularly correct representation of less common behaviours, is important for correct consideration of equity. More precision, particularly temporal precision, is also required for agent-based simulation frameworks such as MATSim \citep{MATSim}.

\subsection{Selected approach}

We seek an approach to activity scheduling that is both practical and accurate. Practicality primarily requires a model that can be estimated on limited data and can be quickly trained and used. Accuracy is the ability of the model to generate populations of schedules with correct distributions; \begin{inline}
    \item individually,
    \item collectively, and 
    \item with respect to their labels
\end{inline}.

We base our model ActVAE on a VAE design. Distinct from existing compositional approaches and GANs, VAEs use an explicit generative approach, directly modelling the target distribution as a probability density so that the real variance of schedules can be better represented. This has previously been demonstrated by \cite{SELF}, although without conditionality.

Distinct from auto-regressive approaches, which are also explicit generative, VAEs use a structured latent, imposing a simplified prior distribution on the generation process. Combined, these allow for both good \emph{density estimation}, \emph{data}, and \emph{ compute efficiency}.

To provide controllable generation with respect to labels we combine regular label injection as per a standard Conditional VAE (CVAE), with a Conditional Prior (CPVAE) using a prior network. The prior network can be thought of as shifting the means and variances of the standard CVAE isotropic prior depending on input labels.

\begin{figure}
    \centering
    \includegraphics[width=0.85\linewidth]{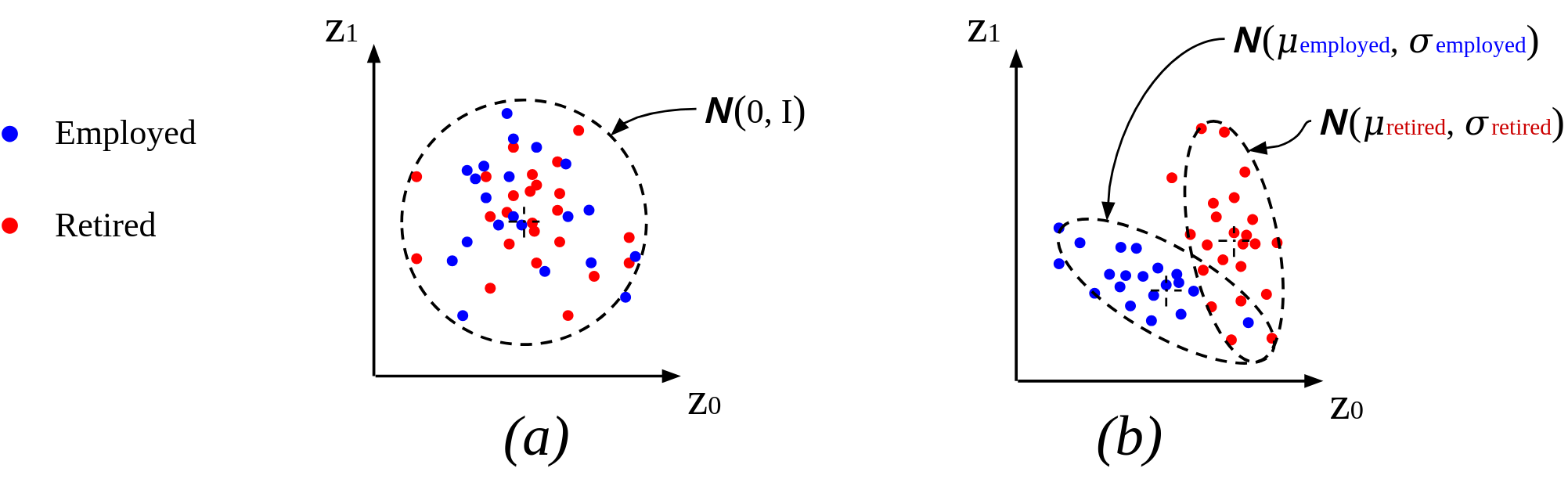}
    \caption{Illustrated latent representations with (a) standard isotropic Gaussian prior, and (b) conditional Gaussian prior.}
    \label{fig:cpvae-latents}
\end{figure}

The intuition behind the conditional prior approach for activity scheduling is that it forces interaction between the conditional and random latents, preventing conditional collapse, but does so in a way that can still allow for noisy information. For example, where a label does not provide much conditional information, its prior distribution can be learned to cover the generality of the latent space with high variance. Where a label does provide information, its prior can be shifted to a specific area of latent space. Similarly, label-conditioned latent representations can overlap, allowing for noise or uncertainty. For example, consider an employed person who usually engages in a work activity, but due to sickness, does not, therefore sharing some part of the latent space with an unemployed or retired person. This concept is illustrated in Figure \ref{fig:cpvae-latents}.

The combination of both regular CVAE label injection with a prior network and conditional prior is supported empirically by ablation in Section \ref{sec:ablations}. The conditional-generative processes of CVAEs and CPVAEs are detailed in \ref{app:vae_deets}.

\section{Methodology}
\label{sec:methodology}

We train our conditional-generative model, ActVAE, to model activity schedules from the UK National Travel Survey (NTS) (Section \ref{sec:data}) and evaluate the resulting model for conditional and generative capabilities as described in Section \ref{sec:eval}. ActVAE is compared to various baseline models detailed in Section \ref{sec:baselines}.

\subsection{Problem definition}
\label{sec:informal_def}

We consider a sample of individuals, each associated with a set of labels ($y$), such as household size, car ownership, and socio-demographic attributes, as well as a recorded activity schedule ($x$). We consider this sample to be drawn from the population, so that it is representative of the conditional distribution of schedules on labels. This data is typically sampled from a real population through a structured travel survey. We aim to model the population probability density of activity schedules, conditional on labels, using this sample, such that the model can then generate new schedules for new samples of input labels.

By modelling the distribution of schedules conditional on labels, we are able to generate schedules dependent on input labels. For example, we can generate a schedule for an individual with a specific set of socio-economic characteristics (potentially previously unobserved). The model can then be used to predict a new distribution of activity schedules for a target distribution of labels. This allows for the model to be used for; 
\begin{inline}
    \item diverse upsampling, including de-biasing, of activity schedule data, and
    \item predicting new activity scheduling behaviour, such as due to demographic shifts or infrastructure and policy interventions. 
\end{inline}

Note that labels do not need to be restricted to socio-economic attributes, but can also include other information potentially relevant to the schedule generation processes discussed in Section \ref{sec:intro}, such as accessibility or weather conditions. Key notation for this section is summarised in Table \ref{tab:methodology_gloss}.

\subsection{Definitions}
\label{sec:definitions}

\subsubsection{Activity schedule definition}

\begin{table}
    \footnotesize
    \caption{Glossary of notation for problem definition in Section \ref{sec:methodology}}
    \vspace{2ex}
    \centering
        \begin{tabular}{l l }
        \hline
        $x$ & schedule \\
        $a_n$ & activity type, indexed in schedule sequence by $n$ \\
        $d_n$ & activity duration, indexed in schedule sequence by $n$ \\
        $L_x$ & schedule sequence length \\
        $T$ & schedules total duration \\
        $y$ & labels composed of multiple variables $y_1, y_2, \dots, y_{N_L}$ \\
        $N_L$ & the total number of label variables \\
        $p(x, y)$ & the supposed \emph{real} joint distribution of schedules and labels \\
        $p(\hat{x}, y)$ & the estimated \emph{real} joint distribution of schedules and labels \\
        $S_\text{real}$ & sample of labels from the \emph{real} distribution with corresponding schedules\\
        $S_\text{target}$ & an abstract sample of labels from some \emph{target} distribution of labels $p(y')$ \\
        $S_\text{eval}$ & labels from $S_\text{real}$, used for generative evaluation \\
        \hline
    \end{tabular}           
    \label{tab:methodology_gloss}
\end{table}

We define a single activity schedule~$x$ as an ordered sequence of activity types~$a_{n}$ with associated durations~$d_{n}$, summing to $T$. As per Section \ref{sec:encoding} travel durations are considered part of activity durations, such that each schedule is composed only of activities. The length $L_x$ of the sequence may vary per schedule, but note we later impose a limit on the maximum number of activities as per Section \ref{sec:encoding}. We choose a total duration of 24 hours, starting and ending at midnight, for our experiments:
\begin{equation}
    x = [(a_{1}, d_{1}), (a_{2}, d_{2}), ... , (a_{L_x}, d_{L_x})], \quad \text{where} \quad \sum_{n=1}^{L_s} d_{n} = T.
\end{equation}

\subsubsection{Labels definition}
\label{sec:labels_def}

We define each schedule as having some associated conditional information, which we refer to as labels $y$. Labels are composed of $N_L$ variables, such as income, gender and so on:
\begin{equation}
    y = [y_1, y_2, \dots, y_{N_L}].
\end{equation}

For simplicity, we represent all variables as nominal tokens, where each token represents membership to a specific category for each label, but future work may wish to experiment with scalar encodings of some variables.

We use labels as a general term to describe any input information that can be used by a model to predict an output schedule. Labels can be composed of any information related to a schedule, such as the day of the week, weather conditions, the age of the individual, their household income, or network information, such as the accessibility of public transport from their home location.

\subsubsection{Real sample definition}
\label{sec:formal_prob}

We consider a sample of pairs of schedules \( x \in \mathcal{X} \), where \( \mathcal{X} \) is the space of all possible schedules, and labels \( y \in \mathcal{Y} \), where \( \mathcal{Y} \) is the space of all possible labels. These pairs of schedules and labels are drawn from a target joint distribution of schedules and labels \( p(x,y) \), such that we consider our sample to have the same joint distribution as the target, which we refer to as the \emph{real distribution}. This distribution describes realised human activity schedules, encompassing both individual preferences and environmental constraints such as accessibility and congestion.

Let the \emph{real sample} be a finite dataset of $N_{real}$ observed schedules and labels:
\begin{equation}
\mathcal{S}_{\text{real}} = [(x_1, y_1), (x_2, y_2), \dots, (x_{N_{real}}, y_{N_{real}})] \subset \mathcal{X}, \mathcal{Y}, \quad \text{with } (x, y) \sim p(x, y).
\end{equation}
For training and monitoring of validation losses, we use an 90\%-10\% split of the real sample. 

\subsubsection{Target labels definition}
\label{sec:target}

For modelling or generating new samples of synthetic schedules, we define a target set of labels:
\begin{equation}
\mathcal{S}_{\text{target}} = \{y_1, y_2, \dots, y_{N_{target}}\} \subset \mathcal{Y}, \quad \text{with } y \sim p'(y).
\end{equation}
Note that these labels can have a novel distribution $p'(y)$ and a variable number of samples $N_\text{target}$. This distribution can be used to represent a change, such as an aged population or better accessibility, that is expected to affect the distribution of schedules.

For our experiments, we use labels from the real sample as the target:
\begin{equation}
\mathcal{S}_{\text{eval}} = \{y_1, y_2, \dots, y_{N_{real}}\} \subset \mathcal{Y}, \quad \text{with } y \sim p(y).
\end{equation}
This allows comparison with the evaluation of the purely generative ActVAE-Generative model, which does not condition on labels

\subsection{Data}
\label{sec:data}

For the real sample of schedules and labels, we extract 59,265 24-hour trip diaries from the 2023 UK National Travel Survey (NTS) trips table\footnote{https://ukdataservice.ac.uk}. Corresponding labels for the real sample are extracted from NTS household and individual data tables. We extract eight labels as detailed in Table \ref{tab:nts-labels}. Labels are ordered by descending Mutual Information (MI) with schedules $I(y_c,x)$. MI measures how much of the variability in NTS schedules can be attributed to each conditioning label. Specifically how much the entropy of schedules is reduced if a label is known. We calculate MI using MINE by \cite{mine}, detailed in \ref{app:mi}.

MI gives an idea of relative importances of each label. However, without an overall idea of the level of entropy of schedules these values are hard to interpret. To give an interpretable idea of label usefulness, we report $trip~rate~\eta^2$ - the proportion of total variance in trip rates explained by each label:
\begin{equation}
    \eta^2 = \frac{SS_{\text{effect}}}{SS_{\text{total}}},
\end{equation}
where $SS_{\text{effect}}$ is the sum of squares attributable to the label of interest, and $SS_{\text{total}}$ is the total sum of squares in trip rates across all observations.

Note that $\eta^2$ values in Table \ref{tab:nts-labels} are low throughout, with labels typically explaining 2\%, or less of the variation of trip rates. Trip rates are important in travel demand models, but only a proxy for considering the overall variance in schedules. We improve upon this analysis using latent representations to estimate combined MI relative to schedules entropy in results Section \ref{sec:latent-label_mi_measurement}.


\begin{table}
    \footnotesize
    \caption{Summary of labels used for experiments}
    \vspace{2ex}
    \centering
    \begin{tabular}{ l l c c }
        \hline
        Label & Categories & Trip rate $\eta^2$ & $I(y_c,x)$* \\
        \hline
        \hline
        Employment & \{ft-employed, pt-employed, education, unemployed, retired\} & 0.0157 & 0.3670 \\ 
        PT accessibility & \{closest, close, mid, far, furthest, unknown\} & 0.0197 & 0.0410 \\
        Day of travel & \{Monday, Tuesday, Wednesday, ... , Sunday\} & 0.0030 & 0.0407 \\
        Household vehicles & \{0, 1, 2, 3, 4, 5, 6\} & 0.0169 & 0.0239 \\
        Household income & \{highest, high, medium, low, lowest\} & 0.0031 & 0.0171 \\
        Household zone & \{urban, suburban, rural\} & 0.0064 & 0.0128 \\
        Sex & \{male, female\} & 0.0005 & 0.0072 \\
        Age & \{0-4, 5-10, 11-15, 16-19, 20-29, 30-39 40-49, 50-69, 70+\} & 0.0224 & 0.0002\\
        \hline
        \multicolumn{4}{l}{}\\[-1em]
        \multicolumn{4}{l}{{\small \textasteriskcentered{ Mean from 5 model runs, variance is $<$0.0005 throughout.}}} \\
    \end{tabular}
    \label{tab:nts-labels}
\end{table}

We select typical individual and household labels used in travel demand models. We additionally include \emph{day of travel} and \emph{transit accessibility} labels to demonstrate the flexibility of the approach for modelling new scenarios. 

\begin{figure}
    \centering
    \includegraphics[trim={8.5cm 0cm 1.75cm 0cm},clip, width=1\linewidth]{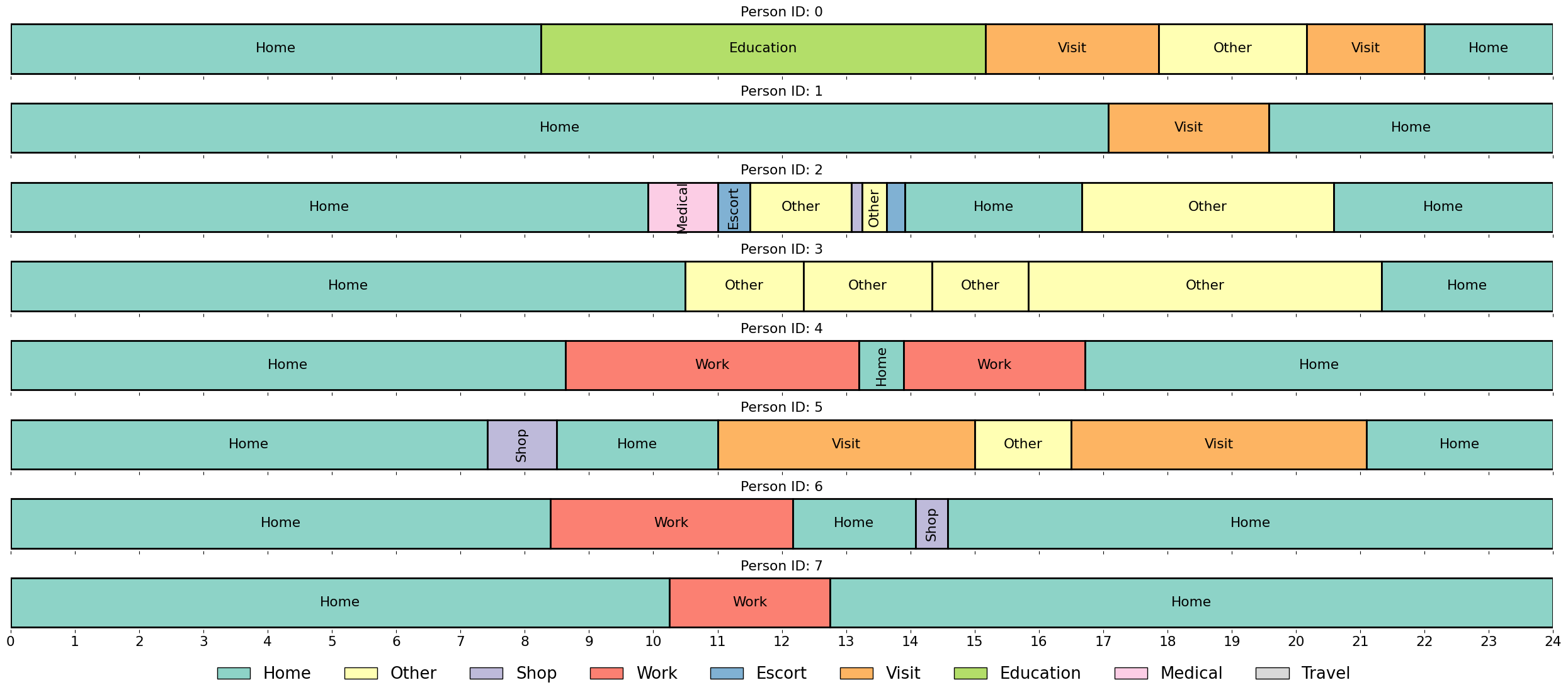}
    \caption{Example UK NTS Schedules}
    \label{fig:nts-examples}
\end{figure}

The NTS trip data is converted into activity schedules using the Foundata\footnote{https://github.com/big-ucl/foundata} project. This process includes:
\begin{inline}
    \item  removing trips, and 
    \item simplifying activity types to the set home, work, education, medical, escort, other, visit, and shop (8 types total). 
\end{inline}
To remove trips, we extend the prior activity durations to include the following trip duration, such that activity start times are maintained. Figure ~\ref{fig:nts-examples} shows examples of extracted and processed activity sequences.

We introduce two types of infeasible schedules:
\begin{inline}
    \item non-home-based - not starting and/or ending at home, and 
    \item consecutive activity - two or more home, work or education activities happening consecutively
\end{inline}. To establish this feasibility in the training set, we filter to exclude non-home-based schedules. Consecutive home, work and education activities are combined into a single activities.

The resulting schedules and labels are then considered as the real sample $\mathcal{S}_{\text{real}}$ and the evaluation sample $\mathcal{S}_{\text{eval}}$, used for generative evaluations. We use a 90\% sample of this data for training and 10\% for validation to monitor over-fitting.

\subsection{Model design}
\label{sec:design}

\begin{figure}
    \centering
    \includegraphics[width=0.75\linewidth]{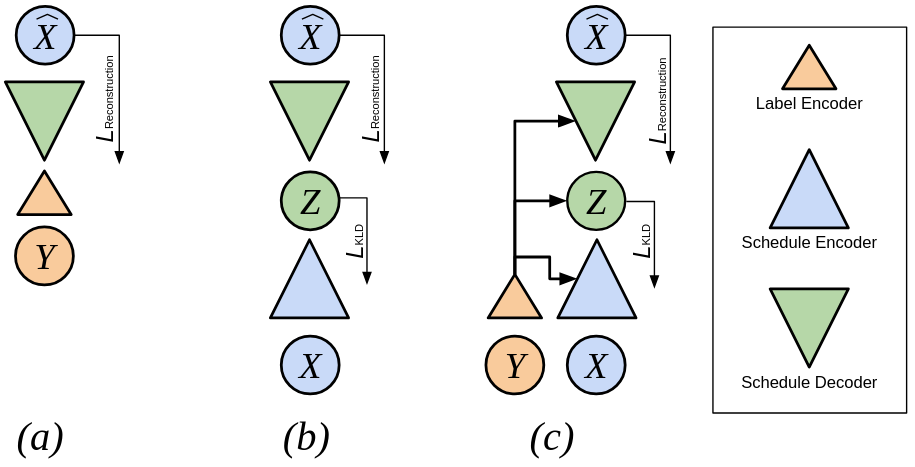}
    \caption{Summary of model encoder and decoder architectures and loss functions (a) ActVAE-Conditional (conditional only), (b) ActVAE-Generative (generative only), and (c) ActVAE (conditional-generative).}
    \label{fig:architectures}
\end{figure}

In this section, we describe the design and operation of ActVAE and its two decompositions: ActVAE-Generative and ActVAE-Conditional. All model variations use consistent components as illustrated in Figure \ref{fig:architectures}, summarised in Table \ref{tab:components}, and detailed in the following sections. Model training and hyperparameters are detailed in Section \ref{sec:training}. Key notation for this section is summarised in Table \ref{tab:model_gloss}. 

\begin{table}
    \footnotesize
    \caption{Summary of components used by ActVAE models}
    \vspace{2ex}
    \centering
    \begin{tabular}{ l l c c c}
        \hline
        Component & Section & ActVAE-Conditional & ActVAE-Generative & ActVAE \\
        \hline
        \hline
        Continuous schedule encoding & \ref{sec:encoding} & \checkmark & \checkmark & \checkmark \\
        Reconstruction loss & \ref{sec:losses} & \checkmark & \checkmark & \checkmark \\
        Schedule encoder & \ref{sec:components} & - & \checkmark & \checkmark \\
        Schedule decoder & \ref{sec:components_decoder} & \checkmark & \checkmark & \checkmark \\
        Conditionality injection & \ref{sec:conditionality_injection} & \checkmark & - & \checkmark \\
        Continuous embedding block & \ref{sec:cont_embed} & \checkmark & \checkmark & \checkmark \\
        Continuous un-embedding block & \ref{sec:cont_unembed} & \checkmark & \checkmark & \checkmark \\
        Labels embedding block & \ref{sec:labels_encoder} & \checkmark & - & \checkmark \\
        Latent block & \ref{sec:latent_block} & - & \checkmark & \checkmark \\
        Prior block & \ref{sec:prior_block} & - & - & \checkmark \\
        \hline
    \end{tabular}
    \label{tab:components}
\end{table}

\begin{table}
    \footnotesize
    \caption{Glossary of model design notation in Section \ref{sec:design}}
    \vspace{2ex}
    \centering
        \begin{tabular}{l l }
        \hline
        $L$ & maximum embedded schedule sequence length (16) \\
        $\mathcal{A}$ & schedule embedding tokens \\
        $N_a$ & number of embedding tokens \\
        $N$ & encoder/decoder block depth (number of layers) \\
        $S$ & schedule embedding hidden size \\
        $H$ & label embedding hidden size \\
        $\alpha$ & loss function duration component weight \\
        $\alpha$ & loss function regularisation component weight \\
        LSTM & Long Short-term memory \citep{LSTM} \\
        $l$ & individual label belonging to labels $y$ \\
        $Y$ & set of individual labels composing $y$ \\
        $c$ & category of label $l$ belonging to set $\mathcal{C_\text{l}}$ \\
        $N_c$ & number of schedules in sample of category $c$ \\
        $N_l$ & number of schedules in sample of category $c$ \\
        $\mathcal{D_\text{c}}$ & category-level joint density evaluation metric \\
        $\mathcal{D_\text{l}}$ & label-level joint density evaluation metric \\
        $\mathcal{D_\text{joint}}$ & domain-level joint density evaluation metric \\
        \hline
    \end{tabular}           
    \label{tab:model_gloss}
\end{table}

\subsubsection{Continuous schedule encoding}
\label{sec:encoding}

The models all use a continuous schedule encoding, where schedules are encoded as sequences of discretely encoded activity types with continuously encoded durations (between zero and one days). This encoding provides a compact and natural representation of schedules and has been shown to outperform the alternative discrete encoding \citep{SELF}. Similar to language models, sequences start with a special \emph{start of sequence} token and are padded as required with \emph{end-of-sequence} tokens, all of zero duration, up to a maximum total sequence length ($L$) of 16. We refer to the combined set of activity tokens (including start and end of sequence padding token) as $\mathcal{A}$ of size $N_a$.

\subsubsection{Continuous reconstruction loss}
\label{sec:losses}

The continuous schedule encoding requires a combined reconstruction loss to incorporate both the discrete representation of activity types and the continuous representation of activity durations. We use categorical cross-entropy for the activity type component of reconstruction loss. Cross-entropy is calculated between activity prediction $p(a)$ and the one-hot encoded ground truth $q(a)$, for each token in $a \in \mathcal{A}$, at each step of the schedule encoding $n$. These are then averaged across all $L$ steps.

For the durations component of the reconstruction loss, we use the mean squared error between the predicted duration $\hat{d}$ and ground truth duration $d$, for each schedule step $n$. We introduce an additional hyperparameter $\alpha$, detailed in Table \ref{tab:hypers}, to weight the duration loss component, giving:
\begin{equation}
\mathcal{L} = - \frac{1}{L} \sum_{n=1}^L \sum_{a \in \mathcal{A}}p(a_n)\log q(a_n) + \frac{\alpha}{L} \sum_{n=1}^L (d_n - \hat{d_n})^2 + \beta D_\text{KL}.
\end{equation}

We additionally employ masking and weighting of the loss function. Firstly, for each sequence, losses are masked to exclude all but the first predicted \emph{end-of-sequence} tokens. This prevents the model from benefiting from the correct prediction of padding tokens. Losses for each sequence are then weighted by the inverse frequency of their labels. Weights are normalised by batch to average one. This weighting biases the model towards schedules with less common label combinations.

\subsubsection{Architecture}
\label{sec:actvae_architecture}

\begin{figure}
    \centering
    \includegraphics[trim={0 0 0 0}, clip, width=0.9\linewidth]{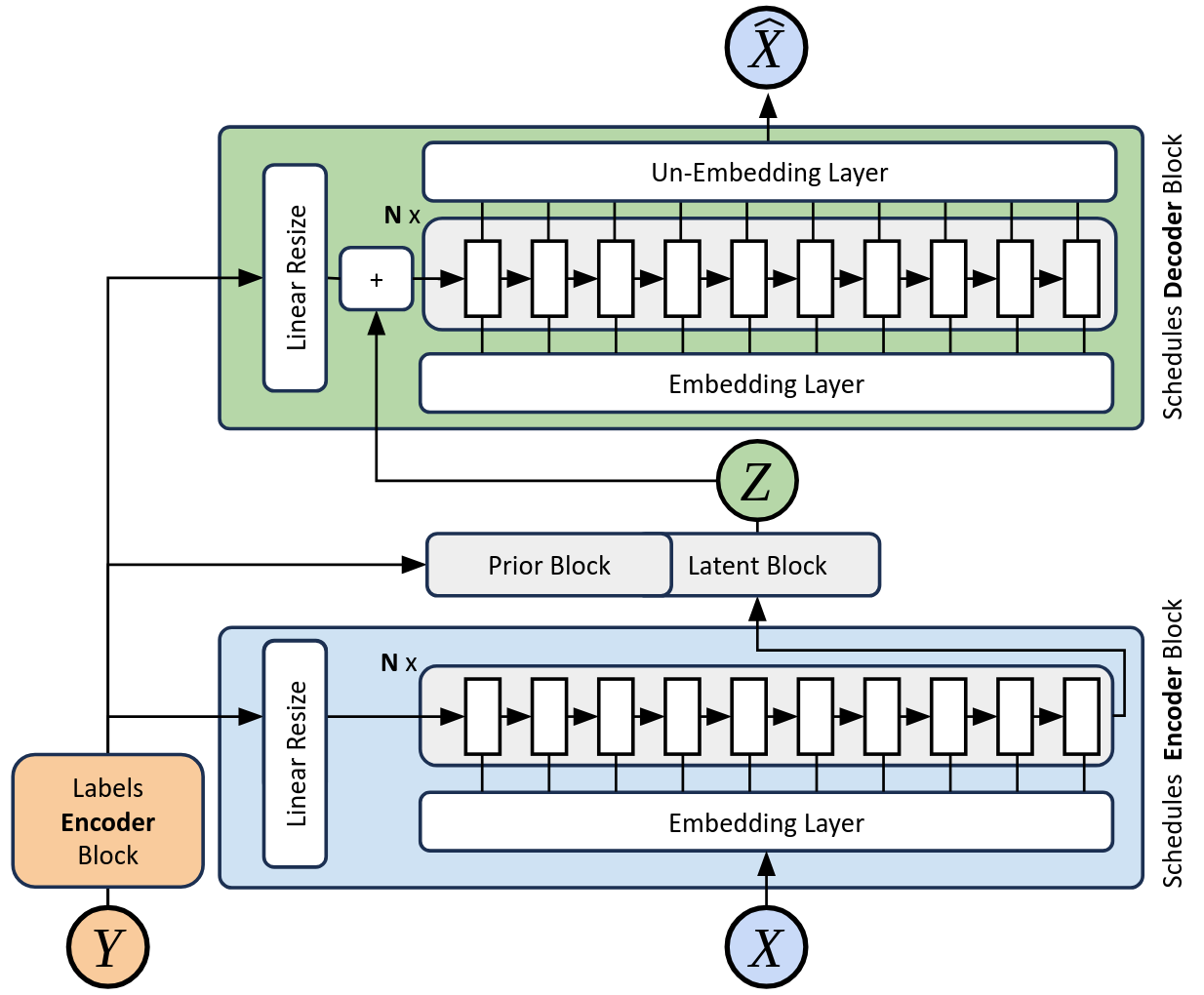}
    \caption{ActVAE architecture}
    \label{fig:cvae}
\end{figure}

The ActVAE architecture  is shown in Figure \ref{fig:cvae}). A brief summary is given here, with a detailed description of the model architecture in \ref{app:actvae}. ActVAE is composed of a schedule encoder, schedule decoder, label encoder, prior block and latent block. 

Input schedules are first embedded as vectors of size $S$ using a bespoke embedding layer. The embedding layer uses learn categorical embeddings for each activity type in a sequence, concatenated with the activity duration. These embeddings are passed iteratively to the encoder composed of $N$ LSTM layers with hidden size $S$. 

The output hidden state of the encoder LSTM block is passed to the latent block to model the hidden latent representation of size six for each schedule.

Latent representations are passed into the initial hidden state of the decoder LSTM block. This is sized with $N$ layers of size $S$ as per the encoder. The output from each decoder LSTM unit is iteratively output as activity type probabilities and durations using an unembedding layer.

\subsubsection{Conditionality}
\label{sec:actvae_conditionality}

Input labels are passed through an encoder block to a latent vector embedding of size $H$. As per a regular CVAE, these are injected into the encoder and decoder architectures, specifically into the initial hidden states of the LSTM blocks, to provide conditionality for generation.

Additionally, as per CPVAE, embedded labels are passed into a prior block, which is used to model label specific latent distributions for the latent representation.

\subsection{Baseline models}
\label{sec:baselines}

We provide three baseline approaches for comparison; \begin{inline}
    \item a traditional compositional approach, \textbf{Compositional}, and 
    \item two ActVAE variations, \textbf{ActVAE-Conditional} and \textbf{ActVAE-Generative}, both based on prior work
\end{inline}. Model capabilities are summarised in Table \ref{tab:experiment}.
\begin{table}
    \footnotesize
    \caption{Summary of ActVAE and baseline model capabilities}
    \vspace{2ex}
    \centering
    \begin{tabular}{ l c c l }
        \hline
        Model Name & Explicit generative  & Conditional & Source \\
        \hline
        \hline
        Compositional & - & \checkmark & Based on \cite{CEMDAP} \\
        ActVAE-Conditional & - & \checkmark & Based on \cite{koushikActivityScheduleModeling2023} \\
        ActVAE-Generative & \checkmark & - & Based on \cite{SELF} \\
        \hline
        ActVAE & \checkmark & \checkmark & This paper\\     
        \hline
        
    \end{tabular}
    \label{tab:experiment}
\end{table}

\subsubsection{Compositional baseline}

We provide a baseline model called \textbf{Compositional} designed to reflect the compositional approach of existing scheduling models as per Table \ref{tab:models}. The model decomposes scheduling into steps composed of discrete and continuous choice models. Models are specified and estimated for each step, and the overall structure is carefully calibrated to provide a fair comparison.

As per CEMDAP \citep{CEMDAP} and DaySim \citep{DaySim}, the first step predicts a daily activity pattern of either; \begin{inline}
    \item \emph{H} - home only,
    \item \emph{W} - work only,
    \item \emph{E} - education only,
    \item \emph{WD} - work plus discretionary,
    \item \emph{ED} - education plus discretionary, and
    \item \emph{D} - discretionary only
\end{inline}. Subsequent steps then predict the number and type of discretionary activities, the duration of activities, and when they occur in the sequence. A final step assembles the previous choices into valid schedules. The design of each component is detailed in \ref{app:compositional}. The code for reproducing these models is available from the Composhed\footnote{https://github.com/big-ucl/composhed} project.

\subsubsection{ActVAE variations}

\begin{figure}
    \centering
    \includegraphics[width=0.75\linewidth]{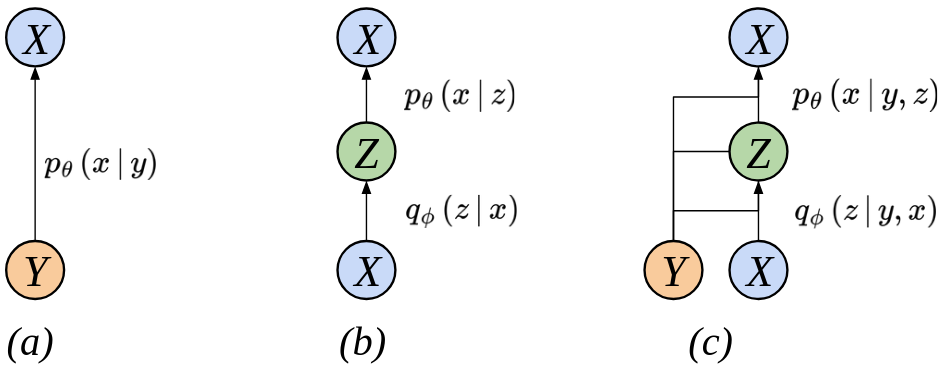}
    \caption{Model generative and conditional processed (a) ActVAE-Conditional (conditional-only), (b) ActVAE-Generative (generative-only), and (c) ActVAE (combined conditional-generative).}
    \label{fig:mods}
\end{figure}

\textbf{ActVAE} combines both conditional and generative processes to generate schedules. The conditional process of ActVAE is analogous to discriminative ML approaches, where variation is modelled depending on input information only. In contrast, the generative process of ActVAE directly tries to model the distribution of schedules as a density distribution. We train and evaluate;
\begin{inline}
    \item \textbf{ActVAE-Conditional}\footnote{Note, the name ActVAE-Conditional is used to indicate that this baseline includes only the conditional components of ActVAE, and not the explicit generative components. ActVAE-Conditional does not include an explicit latent space and so is not a VAE model itself.} - a purely \emph{conditional} model that outputs the most likely schedule given the input labels, based on \cite{koushikActivityScheduleModeling2023}, and
    \item \textbf{ActVAE-Generative} - a purely \emph{explicit generative} VAE without conditional capability as per \cite{SELF}
\end{inline}. These variations are illustrated in Figures \ref{fig:mods} and \ref{fig:architectures}. For comparison, these variations use the same data encodings, model architectures and objectives as ActVAE. The code for reproducing these models is available from the Caveat\footnote{https://github.com/big-ucl/caveat} project.

\subsection{Model training and hyper-parameters}
\label{sec:training}

 We train models until validation loss stabilises, typically for around 100 epochs. We use Adam for gradient descent. Model training hyperparameters are reported in Table~\ref{tab:hypers}. The latent size, $\alpha$ and $\beta$ are selected via an extensive grid search. The remaining hyperparameters are chosen to optimise validation loss using the Tree-structured Parzen Estimator by Optuna \citep{optuna_2019}.

\begin{table}
    \caption{Model hyperparameters}
    \vspace{2ex}
    \centering
    \begin{tabular}{l c c c c c c}
        \hline
        Model & \begin{tabular}{@{}c@{}}Block \\ Size (N.S.H) \end{tabular}  &  \begin{tabular}{@{}c@{}}Latent \\ Size \end{tabular} & \begin{tabular}{@{}c@{}}Learning \\ Rate\end{tabular}  & \begin{tabular}{@{}c@{}}Batch \\ Size\end{tabular}  & $\beta$ & $\alpha$ \\
        \hline
        \hline
        ActVAE & 4.256.64 & 6 & 0.001 & 1024 & 0.01 & 200 \\
        ActVAE-Conditional & 4.128.32 & - & 0.001 & 1024 & - & 200 \\
        ActVAE-Generative & 4.256.- & 6 & 0.001 & 1024 & 0.01 & 200 \\
        \hline
    \end{tabular}
    \label{tab:hypers}
\end{table}

\subsection{Evaluation methodologies}
\label{sec:eval}

For evaluation of each model, after training, a \emph{synthetic sample} of schedules is generated, conditioned on target labels $\mathcal{S}_{\text{eval}}$ as per Section \ref{sec:target}. We refer to the distribution of these schedules as $\hat{p}(x)$ and the joint distribution of schedules and their labels as $\hat{p}(x,y)$.

We do not utilise a partitioned test dataset for our evaluations as we find the high level of uncertainty or variance in label to schedule relationships to make reporting losses uninformative or misleading. In line with other work in this domain, we focus on how well the model can reproduce the distribution of the observed or target samples. We specifically consider the conditional generative capability of the model by considering the distributions of subsamples of schedules based on their conditioning labels. For example, does the distribution of schedules for employed people differ from students and does it differ as per the observed samples? Rather than focus on quantitative evaluation of key metrics only, such as trip frequency, we implement a robust methodology for more broadly estimating the \emph{joint probability density} of labels and their synthetic schedules.

We supplement density estimation with evaluation of model creativity and correctness as defined by \cite{SELF}, label sensitivities, qualitative comparisons, and computational requirements. We additionally use mutual information estimation to quantify the influence of conditional versus random variation for modelling schedules.


\subsection{Density estimation}
\label{sec:density_estimation}

\subsubsection{Schedule density estimation}

Schedule density estimation considers the closeness of the synthetic distribution of schedules $\hat{p}(x)$ to a target distribution of schedules $p(x)$. Density estimation for high-dimensional data is non-trivial, as real and synthetic samples become extremely sparsely distributed. As per \cite{SELF}, we estimate the probability density of schedules by sampling schedules from $\hat{p}(x)$ and measuring a broad range of the resulting marginal distributions, such as the number of trips in each schedule, the number of participations of each activity type, the start times and durations of activity type participations, and so on.

Marginal distributions are then compared pair-wise to distributions measured from the target sample. Differences between distributions are measured using earth-movers distance (EMD). By considering a broad range of marginal distributions, a reasonable approximation of the real combined distributions is reasonably achieved.

For high-level comparison, marginal EMDs are aggregated into the \emph{domain-level} metrics: participations, transitions, and timings. In all cases, lower distances are better, suggesting a closer match between the distribution of real and synthetic schedules.

\subsubsection{Joint schedule-label density estimation}

Schedule density evaluation considers the distribution of synthetic schedules $\hat{p}(x)$ only. We extend this for our generative-conditional case to consider the \emph{joint} distribution of synthetic schedules and their labels $\hat{p}(x, y)$.

\begin{figure}
    \centering
    \includegraphics[width=1\linewidth]{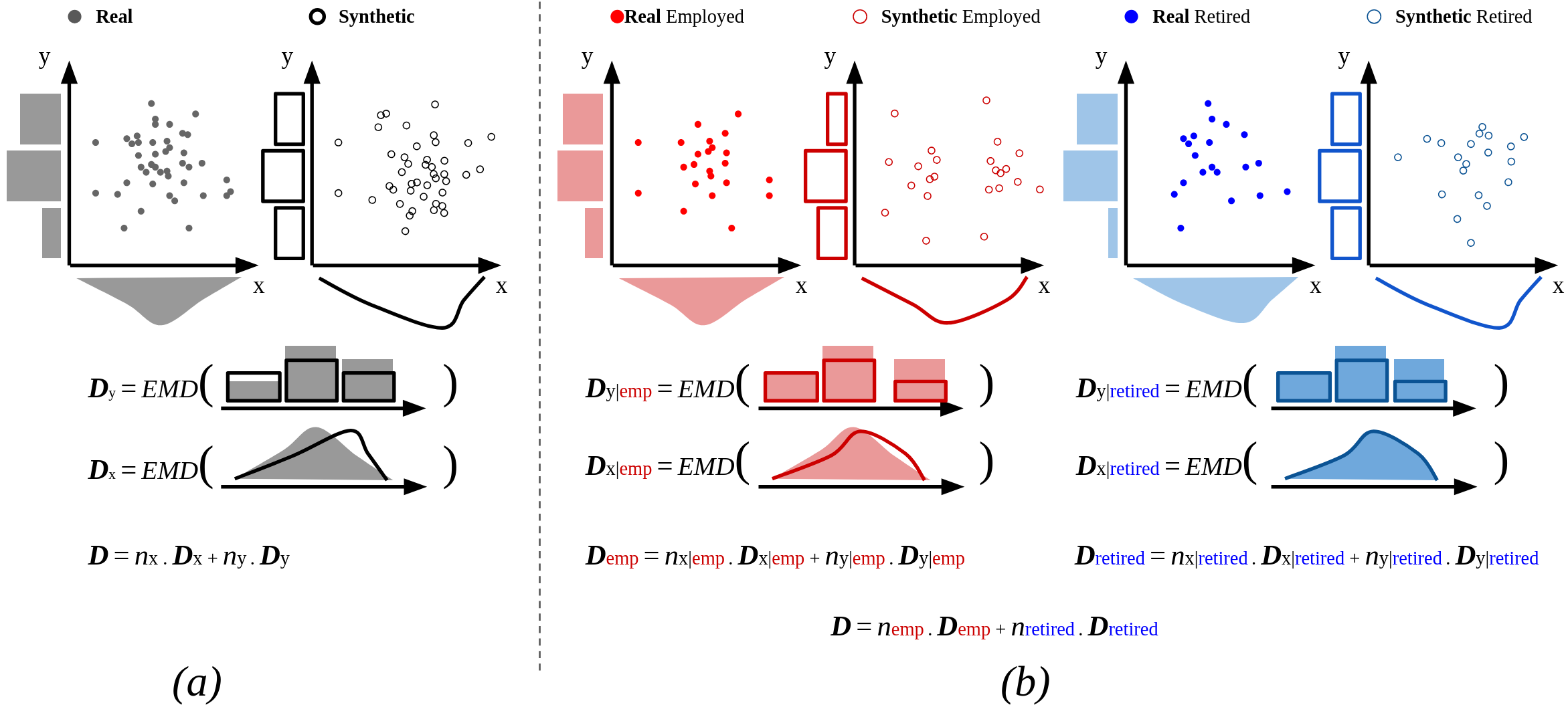}
    \caption{Illustrative comparison of (a) schedule-only density estimation, and (b) joint schedules-labels density estimation. In these examples, the distribution of schedules is imagined as composed of two marginal distributions: a categorical distribution y, and a numeric distribution x. Schedules can be grouped as belonging to either employed persons (red) or retired persons (blue). \textbf{D} denotes a distance metric, EMD is earth-movers distance, and \textit{n} is the number of samples from each marginal, used for weighted sums.}
    \label{fig:density-estimation}
\end{figure}

Similarly to above we further marginalise distributions conditional on label categories, for example, the distribution of number of trips for employed persons is distinguished from the distribution of number of trips for students. Our approach to joint density estimation is illustrated and compared to schedules-only density estimation in Figure \ref{fig:density-estimation}, and detailed in the following. 

For convenience, we define a general density evaluation distance $D_*$ as a function $F$ of real and synthetic samples of schedules:
\begin{equation}
    D_* = F(\mathcal{S_*}, \mathcal{\hat{S}_*}).
\end{equation}

The density estimation of the joint distribution of schedules and labels is approximated by further conditioning on label categories. For example, we can consider the distribution of a sample of schedules conditioned on the \emph{education} employment status, informally, as follows:
\begin{equation}
\mathcal{S}_\text{employed} = \{x_1, x_2, \dots, x_{N_\text{employed}}\}, \quad x \sim p(x \mid \text{\footnotesize employed}).
\end{equation}
This sample is extracted from the real and synthetic samples and used to evaluate density estimation distances conditioned on the \emph{employed} work status:
\begin{equation}
    {D}_\text{employed} = F(\mathcal{S}_\text{employed}, \mathcal{\hat{S}}_\text{employed}).
\end{equation}
This is then repeated for each other work status category (employed, unemployed and so on) and for all other label categories as per Table \ref{tab:nts-labels}. The resulting distances are then averaged, weighted by the size of each sample.

Formally, we consider the combined set of labels $y \in \mathcal{Y}$ as decomposed into individual labels $l \in Y$, where $Y = \{\text{sex}, \text{age}, \text{income}, \text{work status}, \text{zone}, \text{vehicles}, \text{day}, \text{PT access}\}$. Each label then has categories $c \in \mathcal{C}_l$, where $\mathcal{C}_l$ is the possible set of categories for each label $l$.

We then define category-level density evaluation as;
\begin{equation}
    {D}_\text{c} = F(\mathcal{S}_\text{c}, \mathcal{\hat{S}}_\text{c}).
\end{equation}
Label-level joint density evaluation is defined as;
\begin{equation}
\label{eq:label}
    {D}_l = \sum_{c \in C_l} p(y=c) {D}_\text{c}.
\end{equation}
Domain-level joint density evaluation is defined as:
\begin{equation}
    {D}_\text{joint} = \sum_{l \in Y} \frac{1}{|Y|} {D}_l.
\end{equation}

This approach to joint density estimation is limited to \emph{marginal} distributions only. Similarly to the schedule density estimation approach, this prevents consideration of more complex distributions, but by considering all \emph{first-order} label marginals, schedule marginals, and conditionals, a reasonable approximation of the quality of the full joint distribution can be expected.

\section{Results}
\label{sec:results}

Our results are split between \begin{inline}
    \item Section \ref{sec:density_eval} - joint density estimation,
    \item Section \ref{sec:creativity_feasibility} - creativity and feasibility,
    \item Section \ref{sec:label_sensitivity} - label sensitivity,
    \item Section \ref{sec:latent-label_mi_measurement} - estimation of label usefulness,
    \item Section \ref{sec:performance} - model training and generation speed,
    \item Section \ref{sec:case-studies} - \emph{scenarios} to simulate ActVAE application in practice, and
    \item Section \ref{sec:ablations} - an \emph{ablations} study to support model architectural choices
\end{inline}.
\subsection{Joint probability density estimation}
\label{sec:density_eval}

\begin{figure}
    \centering
    \includegraphics[width=1\linewidth]{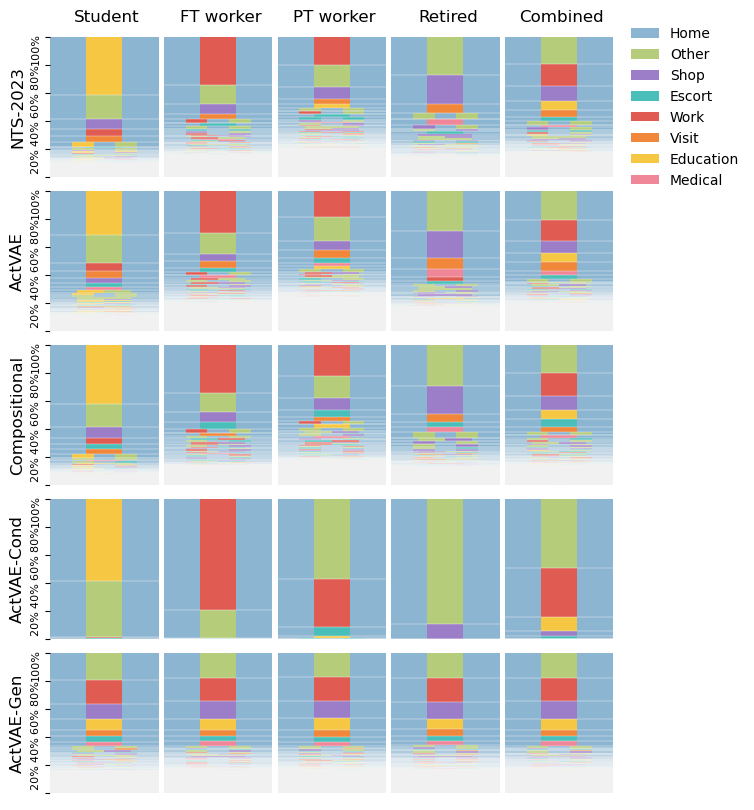}
    \caption{Summary of activity sequence distributions, by employment classes, for (i) the real or target distribution NTS 2023, (ii) ActVAE, (iii) the Compositional baseline model and (iv) ActVAE variations.}
    \label{fig:combined_seq}
\end{figure}

\begin{figure}
    \centering
    \includegraphics[width=1\linewidth]{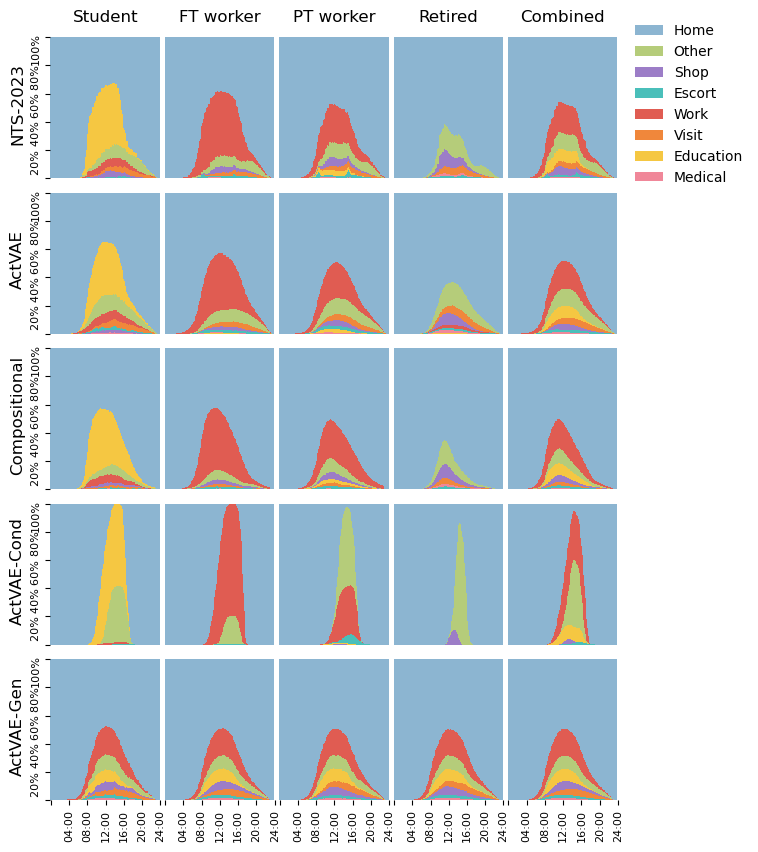}
    \caption{Summary of activity frequency distributions, combined and by employment class, for (i) the real or target distribution, (ii) ActVAE, (iii) the Compositional baseline model and (iv) ActVAE variations.}
    \label{fig:combined_freq}
\end{figure}

Table \ref{tab:eval_summary} presents summary evaluation metrics for joint density estimation (split into the participations, transitions and timing domains). Joint (probability) density estimation considers the similarity between the modelled and target distributions of schedules and their labels. Our goal is for the distributions of generated schedules to closely or exactly match the distributions from the 2023 NTS. This is measured as a earth-movers distance, so that lower is better. We additionally provide qualitative comparisons of activity frequencies (in Figure \ref{fig:combined_freq}) and sequences (in Figure \ref{fig:combined_seq}) for each employment type. 

\begin{table}
    \footnotesize
    \caption{Evaluation summary }
    \vspace{2ex}
    \centering
        \begin{tabular}{l | c | c | c | c | l }
        \hline
          & Compositional & ActVAE-Cond. & ActVAE-Gen. & ActVAE & Unit\\
        \hline \hline
        \multicolumn{6}{l}{}\\[-1em]
        \multicolumn{6}{l}{\textbf{Joint density estimation $\downarrow$}} \\
        \hline
Participations & 0.084 & 0.402 & 0.117 & \textbf{0.083} & rate EMD \\
Transitions & \textbf{0.004} & 0.020 & 0.007 & \textbf{0.004} & rate EMD \\
Timing & 0.080  & 0.200 & 0.057 & \textbf{0.047} & days EMD \\  
        \hline \hline
                \multicolumn{6}{l}{}\\[-1em]
                \multicolumn{6}{l}{\textbf{Creativity $\uparrow$}} \\
                \hline
        Novel & 0.968  & 0.969  & \textbf{0.970}  & 0.969  & prob. \\ 
        Unique & 0.912  & 0.503  & \textbf{0.970}  & 0.967  & prob. \\
        \hline \hline
        \multicolumn{6}{l}{}\\[-1em]
        \multicolumn{6}{l}{\textbf{Feasibility $\uparrow$}} \\
        \hline
Home-based & \textbf{1.000} & 1.000 & 1.000 & 1.000 & prob.\\
None-consecutive** & \textbf{1.000}  & 1.000  & 0.985 & 0.986  & prob.\\
        \hline
        \multicolumn{6}{l}{}\\[-1em]
        \multicolumn{6}{l}{{\small \textasteriskcentered{ Mean from 5 model runs, variance is $<$0.0005 throughout.}}} \\
        \multicolumn{6}{l}{{\small \textasteriskcentered{\textasteriskcentered{ Adjacent home, work or education activities are considered structural zeros.}}}} \\
        \end{tabular}
    \label{tab:eval_summary}
\end{table}

\begin{figure}
    \centering
    \includegraphics[width=1\linewidth]{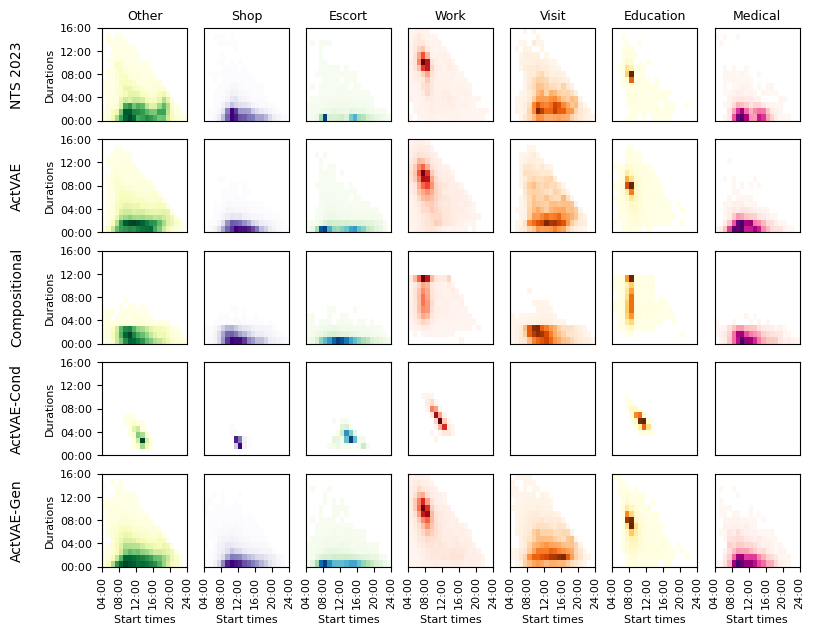}
    \caption{Activity start times and durations}
    \label{fig:timing}
\end{figure}

ActVAE performs best overall at density estimation. For the participations and transitions domains, ActVAE is comparable to the Compositional baseline, but in the timing domain (i.e. when activities happen and for how long) it is significantly better. This is illustrated in Figure \ref{fig:timing}, which plots the joint distributions of start times and durations for each activity type. We see that the compositional approach adds unrealistic artefacts into the distribution of activity start times and durations. Additionally, ActVAE and the generative-only variant better capture the bimodal distribution of escort activities.

The comparably good performance of the Compositional baseline at participations and transitions is likely due to the Daily Activity Plan (DAP)-based structure of the decompositions. DAPs ensure common patterns of participation are generated. This prevents the generation of less common sequences, impacting creativity, but also ensures feasibility. In contrast, ActVAE is not restricted to DAPs, allowing for better creativity and the potential for better density estimation, particularly for less common behaviours that are not captured by DAPs. However, this comes at some cost of feasibility, because the model also has more scope to make mistakes.

The conditional-only variant of ActVAE performs very poorly at density estimation. Without a generative capability, the model is unable to model realistic variation using the input label information. In contrast, the generative-only variant of ActVAE performs well at density estimation, despite not having access to label information. This suggests that the contribution of label information for predicting schedules is minimal.

\subsection{Creativity and feasibility}
\label{sec:creativity_feasibility}

Table \ref{tab:eval_summary} shows high novelty throughout (models rarely generate schedules previously observed in the training data). This suggests that no models are over-fitting or memorising the training data.

Both ActVAE and the generative-only variant are highly creative, generating mostly unique and novel schedules. The Compositional model is similarly novel but less unique. This is caused by simplifications inherent to compositional approaches, which increase the likelihood of duplicate schedules being generated.

The conditional-only variant of ActVAE has no generative capability, so all variance must be driven by changes in labels. For each permutation of labels, the model will generate an estimate of the most likely schedule, leading to significantly lower uniqueness. 

The compositional baseline, ActVAE, and its variants, all guarantee temporally consistent schedules with the desired total durations. In all cases, this ensures the correct use of time as a budget, ensuring activities don't overlap or jump through time.

As defined in Section \ref{sec:data}, we consider a schedule to be feasible if it is both home-based and doesn't containing consecutive home, work or education activities.

A clear advantage of the Compositional model and conditional-only ActVAE variant is that they do not generate infeasible schedules. This is explicitly guaranteed in the DAP-based structure of the Compositional approach, whereas it is learned by the conditional variant and not guaranteed.

ActVAE and the generative-only variant generate feasible schedules approximately 99\% of the time. Their infeasibility is dominated by consecutive home, work and education activities, with none-home-based schedules only very rarely generated. In application, infeasible schedules could be post-processed or rejected with minimal impact on the speed of the approach or distribution.

\subsection{Label sensitivity}
\label{sec:label_sensitivity}

Joint density estimation can be decomposed down to label-level and category-level, allowing comparison of model density estimation capability for individual categories. For example, does the distribution of schedules for employed people in the generated sample look similar to that of employed people in the target sample? Label and category-level distances are shown and briefly described in \ref{app:cat-level}. Distances are presented as EMD, which, although a good measure of distributional shift, is hard to interpret.

To isolate the conditional effect of labels on schedules and show interpretable sensitivity metrics, we consider differences in the expected values of trip rates and other informative schedule features. Considering only the expected values, neglects to consider other distributional shifts, but provides interpretability. In particular, it allows for consideration of the expected change in magnitude. Table \ref{tab:seq_length} presents expected trip rate deltas ($\Delta$) for each label value, i.e. the expected shift in trip rates from the mean, for each label category. These are compared between the target NTS and each model and quantified using MAE so that lower is better.

Both the Compositional baseline and ActVAE capture the expected responses in trip rates observed in the target NTS data. The average MAE of ActVAE is 0.072 trips and Compositional 0.113 trips. The ActvAE conditional-only variant performs poorly with an average MAE of almost a whole trip. The generative-only variant, without conditional capability, does not vary meaningfully, but still performs almost as well as the Compositional baseline with an average MAE of 0.177 trips.

\begin{table}
    \footnotesize
    \caption{Expected trips rates $\Delta$ by label categories, $\Delta$ is difference relative to sample means. }
    \vspace{2ex}
    \centering
        
    \begin{tabular}{l|cc|cc|cc|cc|cc}
    \hline
     &  \multicolumn{2}{ c |}{NTS 2023} & \multicolumn{2}{ c |}{Compositional} & \multicolumn{2}{ c |}{ActVAE-Cond.} & \multicolumn{2}{ c |}{ActVAE-Gen.} & \multicolumn{2}{c}{ActVAE} \\
     Category & Actual & $\Delta$ & $\Delta$* & MAE* & $\Delta$* & MAE* & $\Delta$* & MAE* & $\Delta$* & MAE*  \\
     
    \hline
    \multicolumn{11}{ l }{Age} \\
    \hline
0-10 & 2.670 & -0.261 & -0.296 & \textbf{0.035} & -0.931 & 0.670 & -0.074 & 0.187 & -0.299 & 0.039 \\
10-21 & 2.534 & -0.397 & -0.429 & 0.032 & -0.931 & 0.533 & -0.066 & 0.331 & -0.366 & \textbf{0.031} \\
21-32 & 2.763 & -0.168 & -0.238 & 0.070 & -0.930 & 0.762 & -0.077 & 0.091 & -0.156 & \textbf{0.011} \\
32-40 & 3.072 & 0.141 & 0.024 & 0.117 & -0.904 & 1.045 & -0.065 & 0.207 & 0.090 & \textbf{0.052} \\
40-48 & 3.254 & 0.323 & 0.155 & 0.168 & -0.870 & 1.193 & -0.072 & 0.395 & 0.220 & \textbf{0.103} \\
48-54 & 3.117 & 0.186 & 0.027 & 0.158 & -0.908 & 1.094 & -0.057 & 0.242 & 0.118 & \textbf{0.068} \\
54-61 & 3.089 & 0.158 & 0.023 & 0.135 & -0.923 & 1.081 & -0.049 & 0.207 & 0.113 & \textbf{0.046} \\
61-68 & 3.049 & 0.118 & -0.014 & 0.132 & -0.928 & 1.046 & -0.080 & 0.198 & 0.092 & \textbf{0.026} \\
68-75 & 2.999 & 0.068 & -0.017 & \textbf{0.085} & -0.930 & 0.998 & -0.075 & 0.143 & -0.021 & 0.089 \\
75+ & 2.792 & -0.139 & -0.179 & \textbf{0.041} & -0.930 & 0.792 & -0.062 & 0.077 & -0.209 & 0.070 \\
     
    \hline
    \multicolumn{11}{ l }{Day} \\
    \hline
Monday & 2.909 & -0.022 & -0.127 & 0.105 & -0.916 & 0.894 & -0.067 & \textbf{0.045} & -0.074 & 0.052 \\
Tuesday & 2.946 & 0.015 & -0.097 & 0.113 & -0.917 & 0.932 & -0.063 & 0.078 & -0.031 & \textbf{0.046} \\
Wednesday & 2.993 & 0.063 & -0.049 & 0.112 & -0.911 & 0.973 & -0.075 & 0.138 & -0.006 & \textbf{0.069} \\
Thursday & 2.971 & 0.041 & -0.074 & 0.115 & -0.912 & 0.953 & -0.075 & 0.116 & -0.022 & \textbf{0.063} \\
Friday & 3.018 & 0.088 & -0.010 & 0.097 & -0.915 & 1.002 & -0.064 & 0.152 & 0.003 & \textbf{0.085} \\
Saturday & 2.919 & -0.012 & -0.085 & 0.073 & -0.929 & 0.917 & -0.060 & 0.048 & -0.020 & \textbf{0.008} \\
Sunday & 2.753 & -0.178 & -0.239 & 0.061 & -0.930 & 0.752 & -0.070 & 0.108 & -0.160 & \textbf{0.018} \\
     
    \hline
    \multicolumn{11}{ l }{Employment} \\
    \hline
FT-employed & 2.952 & 0.022 & -0.096 & 0.117 & -0.928 & 0.949 & -0.070 & 0.092 & 0.008 & \textbf{0.013} \\
PT-employed & 3.223 & 0.292 & 0.100 & 0.192 & -0.869 & 1.161 & -0.074 & 0.366 & 0.228 & \textbf{0.064} \\
Retired & 2.948 & 0.017 & -0.044 & \textbf{0.062} & -0.930 & 0.947 & -0.065 & 0.082 & -0.060 & 0.077 \\
Student & 2.525 & -0.405 & -0.463 & 0.057 & -0.929 & 0.523 & -0.072 & 0.333 & -0.389 & \textbf{0.017} \\
Unemployed & 3.130 & 0.199 & 0.051 & \textbf{0.148} & -0.910 & 1.109 & 0.000 & 0.199 & -0.040 & 0.239 \\
Other & 3.240 & 0.309 & 0.171 & 0.138 & -0.839 & 1.148 & -0.048 & 0.357 & 0.186 & \textbf{0.123} \\

    \hline
    \multicolumn{11}{ l }{Household Income} \\
    \hline
Lowest & 2.781 & -0.149 & -0.228 & 0.079 & -0.919 & 0.770 & -0.069 & 0.080 & -0.180 & \textbf{0.030} \\
Low & 2.954 & 0.023 & -0.070 & 0.093 & -0.917 & 0.940 & -0.077 & 0.100 & -0.031 & \textbf{0.054} \\
Mid & 2.933 & 0.003 & -0.086 & 0.088 & -0.921 & 0.923 & -0.057 & 0.060 & -0.026 & \textbf{0.028} \\
High & 3.029 & 0.099 & -0.027 & 0.125 & -0.913 & 1.012 & -0.065 & 0.164 & 0.029 & \textbf{0.070} \\
Highest & 2.956 & 0.025 & -0.073 & 0.098 & -0.922 & 0.947 & -0.070 & 0.095 & -0.011 & \textbf{0.037} \\
     
    \hline
    \multicolumn{11}{ l }{Household Zone} \\
    \hline
Rural & 3.016 & 0.085 & -0.035 & 0.120 & -0.914 & 0.999 & -0.063 & 0.148 & 0.038 & \textbf{0.047} \\
Suburban & 2.765 & -0.165 & -0.234 & 0.068 & -0.927 & 0.762 & -0.065 & 0.100 & -0.179 & \textbf{0.013} \\
Urban & 3.013 & 0.082 & -0.025 & 0.108 & -0.914 & 0.996 & -0.072 & 0.154 & 0.018 & \textbf{0.064} \\

    \hline
    \multicolumn{11}{ l }{PT Access/Egress Distance} \\
    \hline
    Closest & 2.656 & -0.275 & -0.315 & 0.040 & -0.929 & 0.654 & -0.069 & 0.206 & -0.271 & \textbf{0.004} \\
Close & 3.110 & 0.179 & 0.054 & 0.125 & -0.900 & 1.078 & -0.060 & 0.239 & 0.093 & \textbf{0.086} \\
Mid & 3.097 & 0.166 & 0.046 & 0.120 & -0.910 & 1.076 & -0.069 & 0.235 & 0.094 & \textbf{0.073} \\
Far & 3.044 & 0.114 & -0.010 & 0.124 & -0.923 & 1.037 & -0.075 & 0.188 & 0.055 & \textbf{0.058} \\
Furthest & 2.812 & -0.118 & -0.201 & 0.083 & -0.930 & 0.812 & -0.068 & 0.051 & -0.141 & \textbf{0.023} \\
     
    \hline
    \multicolumn{11}{ l }{Sex} \\
    \hline
Female & 2.974 & 0.043 & -0.065 & 0.108 & -0.908 & 0.951 & -0.068 & 0.111 & -0.020 & \textbf{0.063} \\
Male & 2.884 & -0.047 & -0.131 & 0.085 & -0.929 & 0.883 & -0.068 & \textbf{0.021} & -0.070 & 0.023 \\
    
    \hline
    \multicolumn{11}{ l }{Vehicles} \\
    \hline
0 & 2.499 & -0.432 & -0.465 & 0.033 & -0.930 & 0.499 & -0.063 & 0.369 & -0.404 & \textbf{0.028} \\
1 & 2.947 & 0.016 & -0.071 & 0.087 & -0.925 & 0.941 & -0.066 & 0.082 & -0.014 & \textbf{0.031} \\
2 & 3.063 & 0.133 & 0.032 & 0.101 & -0.903 & 1.036 & -0.073 & 0.205 & 0.070 & \textbf{0.062} \\
3 & 3.015 & 0.085 & -0.038 & 0.123 & -0.915 & 0.999 & -0.066 & 0.151 & 0.032 & \textbf{0.052} \\

        \hline \hline
        \multicolumn{11}{l}{}\\[-1em]
      \multicolumn{3}{ l |}{Mean MAE} && 0.113 && 0.931 && 0.177 && \textbf{0.072} \\ 

        \hline
        \multicolumn{11}{l}{}\\[-1em]
        \multicolumn{11}{l}{{\small \textasteriskcentered{ Results from 5 model runs, variance is $<$0.0005 throughout.}}} \\
        \end{tabular}
    \label{tab:seq_length}
\end{table}

\begin{figure}
    \centering
    \includegraphics[trim={0 1.2cm 0 0}, width=1\linewidth]{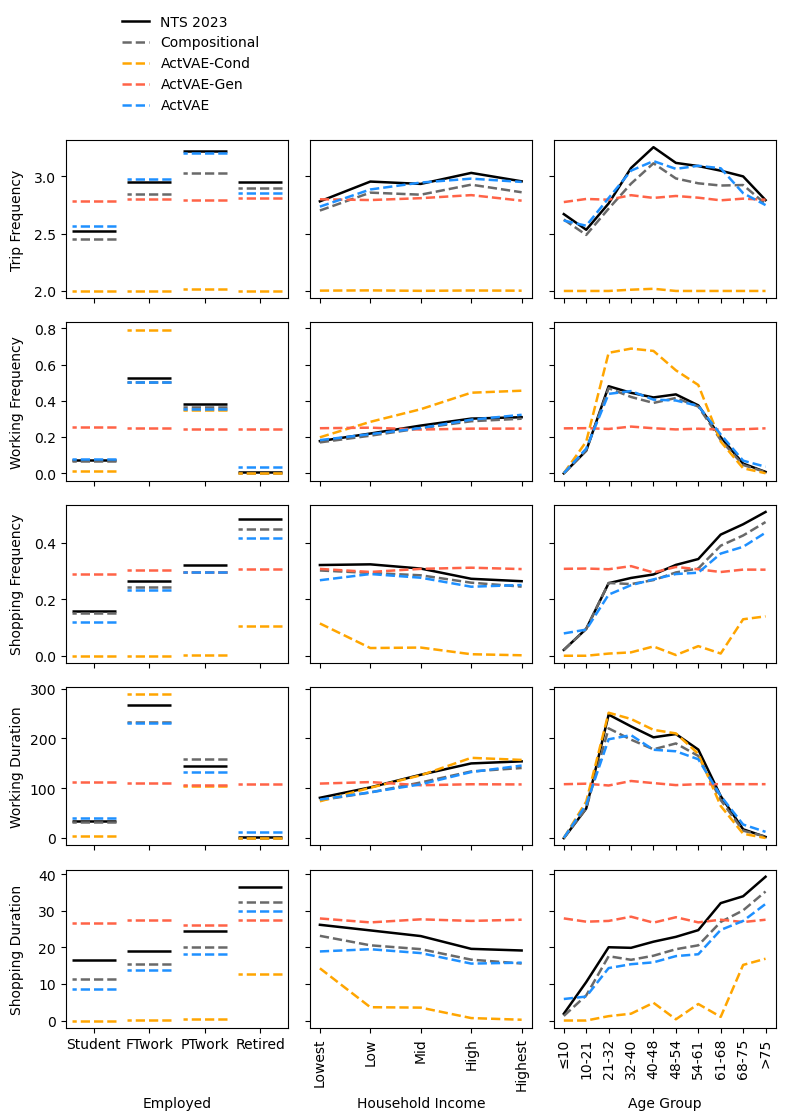}
    \caption{Comparison of expected responses of schedule features to changes in employment, income and age labels.}
    \label{fig:conditional_eval}
\end{figure}

\begin{figure}
    \centering
    \includegraphics[trim={0 1.2cm 0 0}, width=1\linewidth]{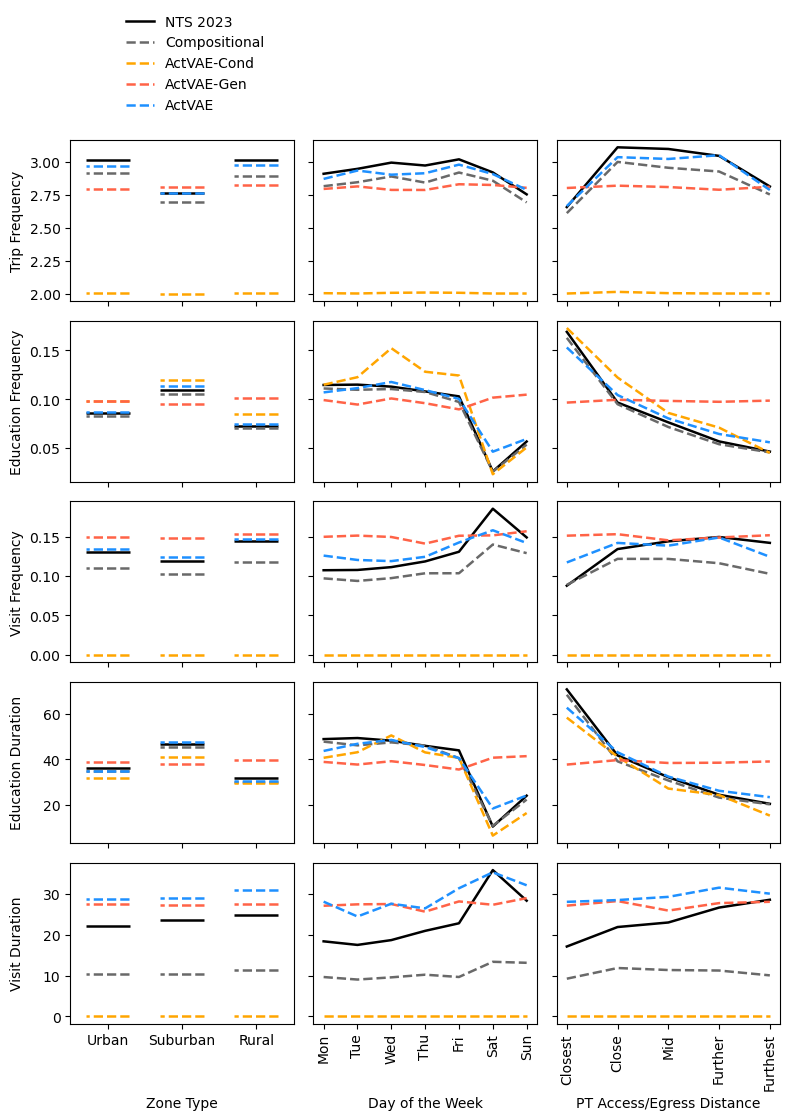}
    \caption{Comparison of expected responses of schedule features to changes in zone type, day and PT access/egress distances.}
    \label{fig:conditional_eval2}
\end{figure}

Trip rates are only a single feature of activity sequences. We present a selection of other features, such as expected work frequencies and shopping durations in Figures \ref{fig:conditional_eval} and \ref{fig:conditional_eval2}. Both ActVAE and the Compositional baseline demonstrate good alignment with the target 2023 NTS expected distributions. Both models show good expected responses to relatively subtle distributions, such as shopping activity durations by income. The conditional-only variant performs unreliably, sometimes demonstrating some expected conditionality, but often not. This is because the information provided by labels is weak and effectively noisy, making it difficult for the model to learn the required conditionality. The generative-only variant has no conditional capability, so it is unresponsive to variations in labels. For a more holistic evaluation of model conditional capability please refer to the category and label-level density estimates in \ref{app:cat-level}.

Overall, the variance in real schedule characteristics, such as trip rates and activity durations, attributable to labels is relatively low. This is described in Section \ref{sec:data} as \emph{weak conditionality} and can be observed in the NTS 2023 trip rate $\Delta$ values in Table \ref{tab:seq_length}. However, ActVAE is able to replicate this weak conditionality well, and reflect the true conditional distribution in the generated schedules. This indicates that ActVAE can be used to forecast the impacts of changes in these variables, for a proposed counterfactual scenario. This is explored explicitly in Section \ref{sec:scenario}.

\subsection{Latent mutual information evaluation}
\label{sec:latent-label_mi_measurement}

\begin{figure}
    \centering
    \includegraphics[width=0.3\linewidth]{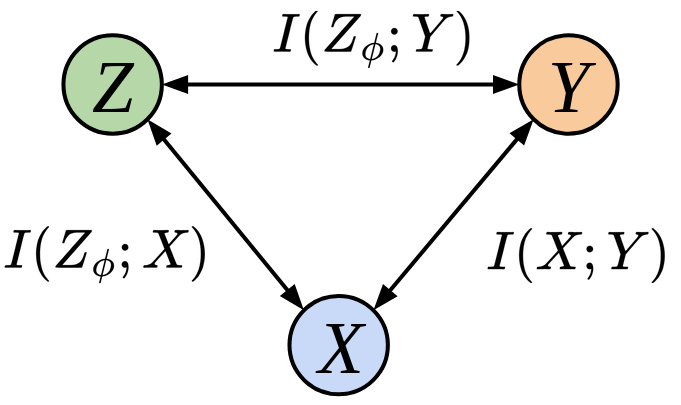}
    \caption{Summary of mutual information estimation, note that $X$ and $Y$ denote real samples of schedule and labels respectively, $Z_\phi$ denotes samples from a latent representation modelled by $\phi$. }
    \label{fig:mi-plan}
\end{figure}

In this section, we question and quantify the importance of conditional versus random variation for modelling schedules using estimates of Mutual Information (MI). MI is a measure of the amount of information shared between two random variables, specifically it is a measure of the reduction in entropy of one variable gained by knowing the other. We use MINE by \cite{mine}, detailed in \ref{app:mi}, for the estimation of MI between \begin{inline}
    \item the real sample of schedules,
    \item real sample of labels, and
    \item latent schedule representations
\end{inline}. We denote the MI between these as; $I(x; y)$, $I(z_\phi; y)$ and $I(z_\phi; x)$ as per Figure \ref{fig:mi-plan}. MI estimates are shown in Table \ref{tab:eval_mi}.

We estimate the reduction in entropy from knowing all labels relative to the entropy of schedules, $I(x;y)/H(x)$. This can be interpreted as the usefulness of labels at explaining the overall variance in schedules.

The entropy of schedules, $H(x)$, cannot be directly measured because it is a high dimensional mixed categorical and continuous representation. Instead, given that;
\begin{equation}
    I(z; x) = H(x) - H(x|z)
\end{equation} and that $H(x|z)$ should be close to 0 for the ActVAE-Generative decoder, we first propose an upper bound as; 
\begin{equation}
    I(x;y)/H(x) \leq I(x;y) / I(z_\phi;x).
\end{equation} 
As per Table \ref{tab:eval_mi}, this upper bound for the relative amount of useful information in the labels is found to be 20\% by the ActVAE-Generative model.

This is also an upper bound because it assumes the models ($\phi$) are able to learn to perfectly embed schedule information in $z$. To correct for this over-estimate we suggest an improved estimate as;
\begin{equation}
    I(x;y)/H(x) \approx I(z_\phi;y) / I(z_\phi;x).
\end{equation}

The intuition is that the information loss in the latent representations is relative for both labels and schedules. As per Table \ref{tab:eval_mi} this improved estimate suggests that only 15\% of schedule variation can be explained using labels. Similar estimates can be made with the ActVAE latent representation, however these estimates are confounded by the introduction of labels to the decoder and latent space.

This analysis confirms earlier results suggesting that the influence of labels is minimal and explains the strong performance of the generative-only ActVAE variant. Essentially, random, or rather unexplained, variance in schedules dominates conditional, or explained variance.

\begin{table}
    \caption{Mutual information estimates; latent-labels $I(z; y)$, latent-schedules $I(z; x)$, and schedules-labels $I(x; y)$}
    \footnotesize
    \vspace{2ex}
    \centering
        \begin{tabular}{l | c c | c c | c c | c | c}
        \hline
           & \multicolumn{2}{c|}{$I(x; y)$ estimate} & \multicolumn{2}{c|}{$I(z_\phi; y)$ estimate} & \multicolumn{2}{c|}{$I(z_\phi; x)$ estimate} & $I(x; y)$ & $I(z_\phi; y)$\\
          Embedding model & mean* & var. & mean* & var. & mean* & var. & $/ I(z_\phi; x)$ & $/ I(z_\phi; x)$ \\
        \hline

-        & 0.528 & 0.002  & - & - & - & - & &  \\

ActVAE-Generative           & - & - & 0.391 & 0.002 & 2.585 & 0.045 & 20\% & 15\% \\
ActVAE                      & - & - & 0.287 & 0.003 & 2.322 & 0.132 & 22\% & 12\% \\

     \hline
        \multicolumn{7}{l}{}\\[-1em]
        \multicolumn{7}{l}{{\small \textasteriskcentered{ Mean estimate from 5 model runs.}}} \\
        \end{tabular}
    \label{tab:eval_mi}
\end{table}

\subsection{Model run times}
\label{sec:performance}

Model training times and generation speeds are assessed using an NVIDIA RTX A5000 GPU. The Compositional baseline, ActVAE and its variations can all be trained in approximately three minutes. However, the Compositional baseline is highly bespoke so required a significant amount of time to assemble and calibrate. In comparison, ActVAE can be rapidly retrained, and if necessary, calibrated using automated hyper-parameter search. This is demonstrated in Section \ref{sec:case-studies} where ActVAE is used for a variety of alternative datasets for similar results.

ActVAE generates 50,000 schedules per second. A full synthetic sample for the UK (68.3 million persons) could therefore be generated in around 20 minutes. The compositional baseline generates 1000 schedules per second and so would be expected to take 50 times longer. This speed makes model and hyperparameter search cheap, moving the bottleneck in model development to the user and the evaluation tools. Alternatively, additional scenarios, or uncertainty, can be more rapidly explored.


\subsection{ActVAE case studies}
\label{sec:case-studies}

\subsubsection{Transit access-egress modelling case-study}
\label{sec:scenario}

We consider the ability of ActVAE to model a withheld sample of schedules and labels. We first define a target/test subsample comprised of schedules for \emph{suburban} zones with the \emph{closest} public transit access-egress distances. This target subsample corresponds to approximately 10\% of the real sample. ActVAE is then trained with this subsample \emph{withheld}. This case-study effectively simulates a scenario where improvements are being made to transit accessibility in suburban areas.

After training, ActVAE is used to generate schedules for the withheld subsample. This experiment tests the ability of ActVAE to generate novel scenarios with previously unseen combinations of labels, i.e. the ability of the model to \emph{extrapolate} outside of the training data distribution.

For evaluation we compare the generated schedules against the withheld target schedules as per Table \ref{tab:scenario_eval_summary}. We additionally provide evaluation of the subsample from the standard model, i.e. with the subsample \emph{included} for training, as a baseline. The scenario and baseline over-estimate the expected number of trips by 0.20 and 0.15 respectively. This is only a minor deterioration caused by the withheld sample. Based on comparable density estimation between the \emph{withheld} and \emph{included} scenarios, ActVAE is able to model the novel scenario relatively well considering the dominance of random variations over conditionality. In some cases the \emph{withheld} scenario performs better than the baseline but this is due to the intrinsic uncertainty in the modelling process. Feasibility is negligibly affected and improvements in creativity are likely a result of noise caused by the out-of-sample generation.

\begin{table}
    \footnotesize
    \caption{Summary of evaluation metrics for the suburban PT access-egress improvements scenario }
    \vspace{2ex}
    \centering
        \begin{tabular}{l | c | c | c | l }
        \hline
          & & Scenario (Subsample  & Baseline (Subsample  & \\
          & Target & \textbf{withheld} from training)* & \textbf{included} in training)* & Unit\\
          \hline \hline
        \multicolumn{5}{l}{}\\[-1em]
        \multicolumn{5}{l}{\textbf{ Trip rates }} \\
        \hline
Expected & 2.60 & 2.80 & 2.74 & trips \\
MAE $\downarrow$ & - & 0.20 & \textbf{0.15} & trips  \\
        \hline \hline
        \multicolumn{5}{l}{}\\[-1em]
        \multicolumn{5}{l}{\textbf{Joint density estimation $\downarrow$}} \\
        \hline
Participations &- & \textbf{0.106} & 0.132 & rate EMD \\
Transitions &- & 0.012 & \textbf{0.006} & rate EMD \\
Timing &- & \textbf{0.040} & 0.058 & days EMD \\  
        \hline \hline
                \multicolumn{5}{l}{}\\[-1em]
                \multicolumn{5}{l}{\textbf{Creativity $\uparrow$}} \\
                \hline
        Novel &- & \textbf{0.999} & 0.970 & prob. \\ 
        Unique &- & \textbf{0.999} & 0.970 & prob. \\
        \hline \hline
        \multicolumn{5}{l}{}\\[-1em]
        \multicolumn{5}{l}{\textbf{Feasibility $\uparrow$}} \\
        \hline
Home-based &- & \textbf{1.000} & \textbf{1.000} & prob.\\
Non-consecutive** &- & \textbf{0.981} & 0.980 & prob.\\
        \hline
        \multicolumn{5}{l}{}\\[-1em]
        \multicolumn{5}{l}{{\small \textasteriskcentered{ Mean from 5 model runs, variance is $<$0.0005 throughout.}}} \\
        \multicolumn{5}{l}{{\small \textasteriskcentered{\textasteriskcentered{ Adjacent home, work or education activities are considered structural zeros.}}}} \\
        \end{tabular}
    \label{tab:scenario_eval_summary}
\end{table}

\subsubsection{Alternative dataset case-studies}

We consider the ability of ActVAE to work with alternative datasets of various sizes from different locations. All datasets are extracted from the Foundata\footnote{https://github.com/big-ucl/foundata} project and use the same pre-processing defined in Section \ref{sec:data} to create new training datasets. For each scenario ActVAE is retrained on the new datasets and then evaluated against them. Model architecture and hyper-parameters are unchanged.

In Table \ref{tab:dataset_eval_summary} we show evaluations are comparable between the 2023 NTS, and three novel datasets of varying sizes from Korea, London and Chicago. The most significant change appears in a drop in feasibility to around 90\% for non-consecutive activities for the smallest dataset, CMAP. This suggests that ActVAE can be used of a variety of scenarios and datasets, although care should be taken for smaller datasets.

\begin{table}
    \footnotesize
    \caption{ActVAE alternative dataset evaluations summary }
    \vspace{2ex}
    \centering
        \begin{tabular}{l | c | c | c | c | l }
        \hline
          & KTDB (Korea) & NTS (UK) & LTDS (London) & CMAP (Chicago) & Unit \\
          \hline
         Training sample size & 89,000 & 57,000 & 37,000 & 18,000 & count\\
                 \hline \hline
        \multicolumn{6}{l}{}\\[-1em]
        \multicolumn{6}{l}{\textbf{Trip rates}} \\
        \hline
Expected & 2.89 & 2.79 & 2.66 & 3.61 & trips \\
MAE$\downarrow$ & \textbf{0.067} & 0.139 & 0.151 & 0.235 & trips \\
        \hline \hline
        \multicolumn{6}{l}{}\\[-1em]
        \multicolumn{6}{l}{\textbf{Joint density estimation $\downarrow$}} \\
        \hline
Participations & \textbf{0.095} & 0.132 & 0.099 & 0.164 & rate EMD \\
Transitions & \textbf{0.006} & \textbf{0.006} & 0.011 & 0.010 & rate EMD \\
Timing & 0.046 & 0.058 & \textbf{0.038} & 0.042 & days EMD \\  
        \hline \hline
                \multicolumn{6}{l}{}\\[-1em]
                \multicolumn{6}{l}{\textbf{Creativity $\uparrow$}} \\
                \hline
        Novel & 0.992 & 0.970 & \textbf{0.997} & \textbf{0.997}  & prob. \\ 
        Unique & 0.979 & 0.970 & 0.996 & \textbf{0.998}  & prob. \\ 
        \hline \hline
        \multicolumn{6}{l}{}\\[-1em]
        \multicolumn{6}{l}{\textbf{Feasibility $\uparrow$}} \\
        \hline
Home-based & \textbf{1.000} & \textbf{1.000} & \textbf{1.000} & 0.998 & prob.\\
None-consecutive** & 0.964 & \textbf{0.986} & 0.983 & 0.899  & prob.\\
        \hline
        \multicolumn{6}{l}{}\\[-1em]
        \multicolumn{6}{l}{{\small \textasteriskcentered{ Mean from 5 model runs, variance is $<$0.0005 throughout.}}} \\
        \multicolumn{6}{l}{{\small \textasteriskcentered{\textasteriskcentered{ Adjacent home, work or education activities are considered structural zeros.}}}} \\
        \end{tabular}
    \label{tab:dataset_eval_summary}
\end{table}

\subsection{Ablations}
\label{sec:ablations}

The contributions of \begin{inline}
    \item the conditional latent prior, and
    \item encoder and decoder label injection\end{inline},
are calculated by removing these components and re-evaluating the resulting ablations. This allows us to evaluate the efficacy of the different elements of the proposed ActVAE structure for capturing the conditional distributions. The results of these ablations are shown in Table \ref{tab:ablation} as percentages compared to the base ActVAE evaluations. Higher values show higher contribution. Note that the latent prior has a more significant contribution to density estimation compared to label injection, but that both are significant and hence our design uses them both.

\begin{table}
\footnotesize
    \caption{ActVAE design ablations. Presented as \% loss in evaluation with feature removed.}
    \vspace{2ex}
    \centering
    \begin{tabular}{l | cc}
    \hline
     & Latent prior & Label injection \\
    \hline
    \hline \hline
        \multicolumn{3}{l}{}\\[-1em]
        \multicolumn{3}{l}{\textbf{Joint density estimation$\uparrow$}} \\
        \hline
participations & 37.3\% & 10.5\%  \\
transitions & 33.8\% & 23.7\%  \\
timing & 19.1\% & 15.3\% \\
        \hline \hline
        \multicolumn{3}{l}{}\\[-1em]
        \multicolumn{3}{l}{\textbf{Feasibility$\uparrow$}} \\
        \hline
        Home based & 0.0\% & -0.0\%  \\
        Consecutive & -0.6\% & 0.6\% \\
        \hline \hline
        \multicolumn{3}{l}{}\\[-1em]
        \multicolumn{3}{l}{\textbf{Creativity $\uparrow$}} \\
        \hline
        Unique & 0.2\% & 0.1\%  \\
        Novel & 0.1\% & 0.1\%  \\  
    \hline
    \multicolumn{3}{l}{{\small \textasteriskcentered{ mean from 5 model runs, variance is negligible.}}} \\
    \end{tabular}
    \label{tab:ablation}
\end{table}


\section{Conclusions}
\label{sec:conclusions}

Based on both quantitative and qualitative evaluation, ActVAE, outperforms the baseline Compositional model. Most notably, ActVAE better models the timing of activities, the diversity of schedules, and manages to capture the bi-modal distribution of escort activities that the compositional baseline does not. Additionally, ActVAE is extremely fast; it does not require manual calibration like traditional compositional approaches and can generate schedules orders of magnitude faster.

We intend for ActVAE to be applied within activity-based demand modelling frameworks for predicting activity participation and timings. In this application, ActVAE has the potential to significantly speed up and simplify new model or scenario development.

Real activity schedules are hugely varied. The impact of this variety on transport demand outcomes is likely non-trivial. Certainly, better representing the diversity of behaviours is important for modelling the equity of outcomes. The capacity of traditional compositional approaches to model this diversity is limited by the assumptions and simplifications made by the decomposed components and their structure. In contrast, ActVAE has the capacity to learn more complex and potentially more realistic features.

\subsection{Limitations}

ActVAE, and more generally, deep generative ML approaches, present trade-offs that should be carefully considered before application in broader modelling frameworks. The first of these is a feasibility-creativity trade-off. ActVAE allows for more creativity, resulting in more diverse outputs and, therefore, the potential for better density estimation, but this comes at the cost of some infeasible outputs.

Note that ActVAE guarantees temporal consistency, preventing overlaps or gaps in time budgets. Our definitions of infeasibility are that a schedule must \begin{inline}
    \item be home-based (start and end with a home activity), and
    \item not contain consecutive home, work or education activities
\end{inline}. These are artificial constraints based on existing model limitations. The compositional model, for example, cannot generate non-home-based schedules. But non-home-based schedules do exist in the real sample. We therefore suggest interpreting the poorer feasibility of ActVAE as a limitation on model \emph{compliance}. Using a compositional framework, a modeller can usually impose assumptions of schedule structure by manually specifying and calibrating model components and their structure. ActVAE must implicitly learn the required structure from data and is prone to mistakes around 1\% of the time. However, compliance can be simply and cheaply achieved with ActVAE using rejection sampling.

Secondly, unlike theory-based compositional approaches, ActVAE is a ``black~box'' and therefore not intrinsically \emph{interpretable} by examination of its parameters. In fact, ActVAE model sensitivities can be cheaply explored to get high confidence in model behaviour, which we demonstrate with a trip rate sensitivity analysis. However, this \emph{explainability} does not facilitate any controllability of the model by the modeller. In contrast, theory-based modellers can calibrate baseline scenarios and simulate new scenarios by interpreting and directly modifying parameters. Such \emph{extrapolation} of model parameters beyond the data is not feasible with ActVAE; instead, the model must rely on information in its training data, which we can think of as \emph{interpolation}.

The inability to \emph{extrapolate} outside observed data limits the application of ActVAE to near-term scenarios or where interpolation from existing behaviours is reasonable, for example, from an ageing population, or variations in accessibility (as demonstrated in the experiment in Section \ref{sec:scenario}). For scenarios that wish to explore more novel changes to scheduling behaviours, for example, radical changes in preferences due to new infrastructure, other or additional approaches would likely be required.

\subsection{On deep generative models for human behaviour more generally}

We show that a generative-only variant of ActVAE is able to perform comparably to other models, despite not using conditional information to model schedules. This suggests that the influence of unobserved or unobservable processes dominates activity scheduling. This assertion is quantified in Section \ref{sec:latent-label_mi_measurement}, where we use mutual information estimation to show that only around 15\% of the information in the latent representation of schedules can be attributed to explanatory variables.

Traditional compositional approaches are structured around discriminative discrete choice models. Unexplained and unexplainable variation is therefore captured in the error terms of these models and their interactions accumulating through subsequent steps. The ability of such models to capture realistic variation is therefore limited by the ability of the modeller to manually structure and calibrate these interactions. In contrast, deep generative approaches can explicitly model the required random variation.

This finding is likely significant beyond scheduling. Humans are complex, and so our behaviours are complex. Much of the rationale, if there is some, or causation for a particular decision or process, can reasonably be assumed to be unobserved or unobservable by the modeller. Variation from such unobserved and unobservable processes must therefore be captured by random processes in our models. Where behaviours are complex and explanatory data relatively limited or weak, deep generative approaches to modelling this behaviour are likely useful.

\subsection{Further Work}
\label{sec:further}

Deep generative modelling is an enormous domain, and approaches to human activity schedule generation have only begun to be explored. We share some experiments to support the design of ActVAE, but anticipate that other approaches and architectures may be found that perform better, particularly in scenarios with much higher data availability.

Our evaluation approach is a significant departure from other research, which tend to focus on a limited number of key output metrics, such as trip rates. Instead, we estimate and compare the joint distribution of labels and schedules more generally. This new focus towards distribution should be welcome, especially where the representation of the real diversity of human behaviours is important. However, our approach does not allow decomposition of the generative versus conditional quality of models. As such, we find the quantitative evaluation of the \emph{conditional} capability of models challenging. A general quantitative conditional evaluation process would be welcome. A useful avenue of progress would be the adoption of supplementary classifier models, similar to the Fréchet Inception Distance (FID) and Inception Score (IS) paradigms for image generation.

Our existing work is limited to 24-hour activity scheduling of individuals. Further work should progress towards a more complete approach to activity-based modelling by incorporating trips, distances, locations and modes. Additionally, future work could explore the approach for jointly modelling households and longer durations.

\appendix

\section{Methodology details}
\label{app:vae_deets}

In this section, we detail the technical formulation of ActVAE, which combines a regular Conditional VAE (CVAE) architecture with a Conditional Prior VAE (CPVAE). Key notation for this section is summarised in Table~\ref{tab:vae_gloss}.

\begin{table}
    \footnotesize
    \caption{Glossary of VAE notation for Section \ref{app:vae_deets}}
    \centering
        \begin{tabular}{l l }
        \hline
        $x$ & schedule \\
        $y$ & labels (conditional data) \\
        $z$ & a latent embedding or representation \\
        $p(x)$ & an observed probability distribution of schedules \\
        $p(z)$ & the latent prior distribution \\
        $p(\hat{x})$ & an estimated probability distribution of schedules \\
        $q_\phi(z \mid x)$ & VAE encoder block, also \emph{inference} process \\
        $p_\theta(x \mid z)$ & VAE decoder block, also \emph{generative} process \\
        $q_\phi(z \mid x, y)$ & conditional VAE encoder block, also \emph{inference} process \\
        $p_\theta(x \mid z, y)$ & conditional VAE decoder bloc, also \emph{generative} process \\
        $\mathcal{L}$ & loss function \\
        $D_\text{KL}$ & Kullback-Leibler divergence \\
        $\mathcal{I}$ & mutual information \\
        \hline
    \end{tabular}           
    \label{tab:vae_gloss}
\end{table}

\subsection{Conditional VAEs}
\label{sec:cvae}

ActVAE is based on the Conditional VAE (CVAE) architecture, by \cite{kingma_cvae}. CVAEs extend the classic VAE architecture to \emph{inject} labels so that the decoder can learn the conditional distribution $p(x \mid y)$. Where for this work, $x$ are schedules and $y$ their associated explanatory labels, such as age, income, and so on.

The \emph{generative process} \(p_\theta(x \mid y)\) is modelled as conditional on $y$ and a random latent variable $z$:
\begin{equation}
p_\theta(x \mid y) = \int p_\theta(x \mid z, y)\, p(z)\, dz.
\end{equation}

To correctly map $p(x)$ across the latent space, and therefore learn inter-sample random variation, CVAEs additionally learn an \emph{inference process}, \( q_\phi(z \mid x, y) \), parameterised by a neural network called the \emph{encoder}.

The encoder provides an auxiliary mapping from input schedules and labels to the latent $z$. This allows formulation of the Evidence Lower Bound (ELBO) $\mathcal{L}$, which provides a tractable objective for training that lower-bounds the log-likelihood of schedules given labels:
\begin{equation}
\label{eq:elbo}
\mathcal{L}(\theta, \phi; x, y) =
\mathbb{E}_{q_\phi(z \mid x,y)} \left[ \log p_\theta(x \mid z,y) \right]
- \mathrm{D}_{KL}\big(q_\phi(z \mid x,y) \,\|\, p(z)\big).
\end{equation}
The first term encourages accurate reconstruction of the data, training the decoder to map latent samples to realistic outputs. The second term regularises the encoder by penalising divergence from the prior distribution of $z$, ensuring that the latent space remains structured and supports random sampling from the prior distribution. For training, we use an activity schedule specific generalisation of the ELBO, detailed in Section \ref{sec:losses}.

\subsection{Disentangling latent spaces}

When correctly trained, the CVAE decoder is able to generate schedules conditional on input labels and on random samples from the latent prior. These two processes can be considered as capturing \begin{inline}
    \item deterministic variation in scheduling, for example, due to varying employment status or age, and
    \item random variation in scheduling, for example, due to unobserved factors, such as remembering to set an alarm clock.
\end{inline}

It is useful to think of both the label $y$ and the random latent $z$ as combining to form a joint latent representation of each schedule. For controllable generation, i.e. where new samples can be generated for specific labels, the known and random components of the combined latents need to be disentangled such that they are independent. Where $z$ and $y$ are dependent, CVAEs are prone to conditional collapse, where all information for label reconstruction is passed through the random latent, ignoring the input labels.

\subsection{Conditional Prior VAE}
\label{sec:cpvae}

A typical CVAE uses a standard isotropic Gaussian prior for $z$ with fixed standard $\mu$ and $\sigma$ for all samples of $x$ and $y$. A Conditional Prior VAE (CPVAE) makes the prior more expressive by learning a prior network $z = \psi(y)$, such that the generative process becomes:
\begin{equation}
p_\theta(x \mid y) = \int p_\theta(x \mid z, y)\, p_\psi(z \mid y)\, dz, \quad \text{with } p_\psi(z \mid y) = \mathcal{N}(\mu_\psi(y), \Sigma_\psi(y))
\end{equation}

Note that we still model the prior distribution as Gaussian, but make $\mu$ and $\sigma$ conditional on labels. This allows the $\mathrm{D}_{KL}$ term from the objective (Equation \ref{eq:elbo}) to remain tractable.

\section{ActVAE Model Design}
\label{app:actvae}

\subsection{Schedule encoder}
\label{sec:components}

The ActVAE architecture (Figure \ref{fig:cvae}) is composed of an encoder block and a decoder block. As per Section \ref{sec:cvae}, the encoder parametrises the inference process $p_\phi(z \mid x, y)$, the decoder the generation process $p_\theta(x \mid z, y)$. We use a latent block detailed in Section \ref{sec:latent_block} to parametrise the prior $p(z)$.

The encoder block iteratively progresses through each step of the input sequences, first embedding input activity type and duration into a size $S$ hidden vector as per Section \ref{sec:cont_embed}, then passing into $N$ stacked LSTM units of size $S$. The LSTM units pass a learnt hidden state from step to step. The output hidden state from the final unit is flattened and passed via a feed-forward block, with output size $S$, to the latent block.

\subsection{Schedule decoder}
\label{sec:components_decoder}

The decoder block is similarly based around $N$ stacked LSTM units of size $S$. The size $S$ and depth $N$ control the complexity or capacity of the hidden states of the encoder and decoder blocks. The output from the latent block is first resized using a feed-forward block and reshaped into the hidden state of the first decoding LSTM unit. This first unit uses a start-of-sequence token as input, further steps then use the argmax sampled output from the previous step. During training, we use 50\% teacher forcing, where previous predictions are replaced by training data, to improve training stability. Input sequence steps are embedded as per the encoder. LSTM unit outputs are un-embedded from size $S$ vectors into predicted activity types and durations using un-embedding blocks as per Section \ref{sec:cont_unembed}.

\subsection{Conditionality injection}
\label{sec:conditionality_injection}

To provide label context for conditionality, labels are first encoded using a label encoder block (Section \ref{sec:labels_encoder}) to output a hidden vector of size $H$. This vector representation of labels is then input to the encoder and decoder block, as per Figure \ref{fig:cvae}. After resizing via linear layers, the label hidden state is input into the encoder initial LSTM unit hidden state and added to the decoder initial hidden state.

\subsection{Embedding layer}
\label{sec:cont_embed}

The continuous schedule encoding combines a discrete activity type encoding and continuous duration encoding. This encoding is embedded into a vector of size $S$ using a custom \emph{continuous embedding layer}. Input activity types and durations are first split, and then tokens are transformed into vectors of size $S-1$ using a learnt embedding layer. The resulting activity embedding vectors are then concatenated back with their durations. Note that activity types also include the special start and end of sequence tokens.

\subsection{Un-embedding layer}
\label{sec:cont_unembed}

At each step, the model LSTM blocks output vectors of size $S$. These are unembedded into activity type probabilities and durations using a custom \emph{continuous un-embedding layer}. Output vectors are first resized using a fully connected linear layer, which outputs vectors of size $N_a + 1$ (number of token types, including special start and end of sequence tokens, plus a duration scalar). Each duration is passed through a sigmoid activation so that all duration values are between zero and one. The remaining activity type vectors are decoded into probabilities using a softmax layer. 

Completed sequence durations are then normalised by the total duration of each predicted sequence. This normalisation enforces the intended duration $T$. Note that the calculation of total predicted duration excludes any durations \emph{after} the first end of sequence token.

For generation, we sample from activity type probabilities using argmax. Activities and their durations for each sequence are then concatenated back together. Argmax sampling removes stochasticity from the architecture, forcing all variance to be modelled by the VAE generative process.

\subsection{Label encoder block}
\label{sec:labels_encoder}

\begin{figure}
    \centering
    \includegraphics[width=.7\linewidth]{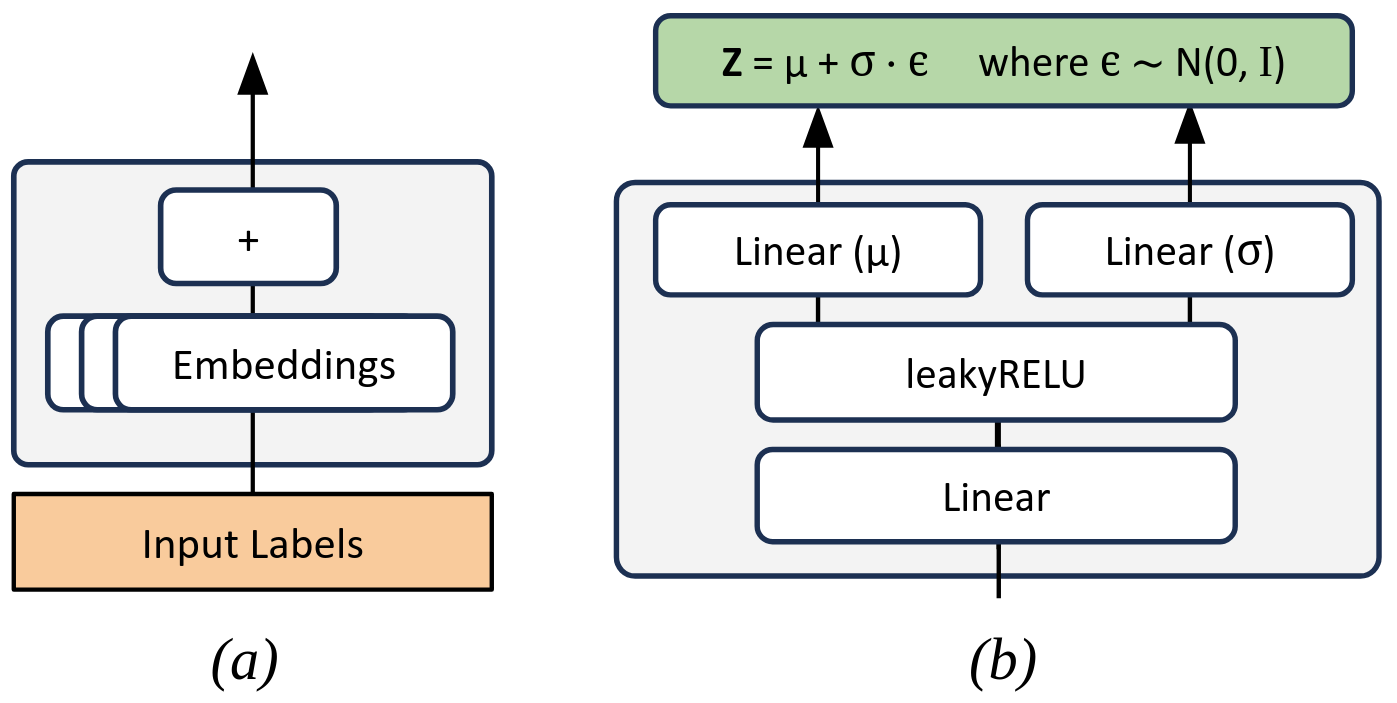}
    \caption{(a) Label encoder block, and (b) latent block}
    \label{fig:blocks}
\end{figure}

The label encoder block in Figure \ref{fig:blocks}(a) is used to embed multiple categorical labels as a vector of size $H$. The label encoder block first individually encodes each categorical label using learnt embeddings of size $H$. The resulting vectors are combined through addition.

\subsection{Latent block}
\label{sec:latent_block}

Figure \ref{fig:blocks}(b) presents the latent block design. The latent block input is the encoded schedule representation from the schedule encoder block. This hidden schedule state is first passed through a linear layer and a LeakyRELU layer, both of size $S$. The block then uses two linear layers to output $\mu$ and $\sigma$ vectors, both of size six. The latent vector $Z$ is sampled from these means and variances using an ancillary randomly sampled vector $\epsilon \sim \mathcal{N}(0, I)$. This is commonly referred to as the \emph{reparametrisation trick}.

\subsection{Prior block}
\label{sec:prior_block}

ActVAE additionally uses a prior block. This is as per the latent block in Figure \ref{fig:blocks}, but receives as input the label hidden state of size $H$. The prior block also outputs $\mu$ and $\sigma$ vectors, both of size six. During training, these vectors are treated as learnt priors $\mu_{prior}$ and $\sigma_{prior}$ for each input combination of labels. These label-specific priors are passed to the $\mathcal{D}_{KL}$ loss component to regularise the encoder latent representation of each schedule depending on its labels. After training, the latent block is discarded, and instead, $\mu_{prior}$ and $\sigma_{prior}$ are used to generate new latent representations using the reparametrisation trick.

\section{Compositional Model Design}
\label{app:compositional}

Following the decomposed and econometric tradition of activity-based modelling as per ActivitySim \citep{activitysim}, CEMPDAP \citep{CEMDAP} and DaySim \citep{DaySim}. The compositional baseline model generates 24-hour activity schedules through a sequential hierarchy of five independently estimated statistical models and an assembly step.

\textbf{Step 1 - Daily Activity Pattern}

A multinomial logit model is used to classify Daily Activity Patterns (DAPs) into one of six types; \begin{inline}
    \item \emph{H} - home only,
    \item \emph{W} - work only,
    \item \emph{E} - education only,
    \item \emph{WD} - work plus discretionary,
    \item \emph{ED} - education plus discretionary, and
    \item \emph{D} - discretionary only,
\end{inline}. The DAP classification is conditioned on input labels and determines which subsequent steps are executed.

\textbf{Step 2 - Mandatory activity duration}

For individuals assigned a \emph{W}, \emph{E}, \emph{WD} or \emph{ED} DAPs, the duration of the work or education activity is sampled from a log-normal regression model conditioned on input labels and DAP type. Stochastic noise is added at generation time by sampling from the residual distribution of the fitted model.

\textbf{Step 3 - Number of discretionary activities}

For individuals assigned a \emph{WD}, \emph{ED} or \emph{D} DAPs, ordered logit models predict the number of discretionary activities of each type (shop, visit, escort, medical, other), conditioned on input labels and the remaining time budget after mandatory activity duration allocation.

\textbf{Step 4 - Discretionary activity types}

For each discretionary activity slot (first, second and third or more occurrences), a multinomial logit model predicts the activity type from the set {shop, visit, escort, medical, other}, conditioned on input labels, DAP, slot and remaining time budget.

\textbf{Step 5 - Discretionary activity durations}

Log-normal regression models predict the duration of each discretionary activity, estimated separately per activity type and conditioned on input labels and the remaining time budget at that point in the sequence. Stochastic noise is added at generation time by sampling from the residual distribution of the fitted model.

\textbf{Step 6 - Schedule assembly}

For assembly, we use a rule-based algorithm to assemble outputs from previous components into a valid 24-hour schedule. For each employment category, mandatory activity start times are sampled from a kernel density estimate fitted to empirical times. DAP types \emph{W} and \emph{E} schedules are then assembled from the activity start time and duration. Duration is clipped to ensure at least 30 minutes of home activity at the end of the schedule.

For the combined mandatory and discretionary DAPs \emph{WD} and \emph{ED}, a logistic regression, conditioned on the labels, mandatory start time and activity type is used to assign each discretionary activity to either; \begin{inline}
    \item the mandatory tour - pre-mandatory activity,
    \item the mandatory tour - post-mandatory activity,
    \item a new tour - pre-mandatory activity, and
    \item a new tour - post-mandatory activity    
\end{inline}.

Discretionary activities and additional home activities are then placed as per their start times and durations, but with the following enforcement: \begin{inline}
    \item remove overlaps,
    \item minimum home duration of 10 minutes, and
    \item where there is not enough budget, discretionary durations are scaled.
\end{inline}

For discretionary-only DAPs, the first start time is sampled from a kernel density estimate fitted to empirical times. A logistic regression, conditioned on the labels, slot number and remaining budget, then predicts if each following activity should be in the same home-based tour, or start a new one. Activities are then assembled using the same enforcement above.

\section{Mutual information estimation methodology}
\label{app:mi}

We use MINE by \cite{mine} for the estimation of mutual information (MI) between the real sample of schedules, labels, and inferred or learnt samples of latent representations as per Figure \ref{fig:mi-plan}. MINE estimates MI between two random variables using the Donsker–Varadhan representation of the Kullback–Leibler divergence:
\begin{equation}
I(A; B) = \sup_{T \in \mathcal{T}} \mathbb{E}_{p(a,b)}[T] - \log \left( \mathbb{E}_{p(a)p(b)}[e^T] \right),   
\end{equation}
where $T$ is an \emph{optimal critic} able to distinguish between dependent and independent pairs, sometimes called the \emph{statistics} network or model. $T$ is approximated using a neural network trained to assign high scores to samples drawn from the joint distribution $p(a,b)$, and low scores to samples from the product of marginals $p(a)p(b)$, which represent independent pairs. We estimate marginals throughout by simply shuffling inputs for each batch.

\begin{figure}
    \centering
    \includegraphics[width=1\linewidth]{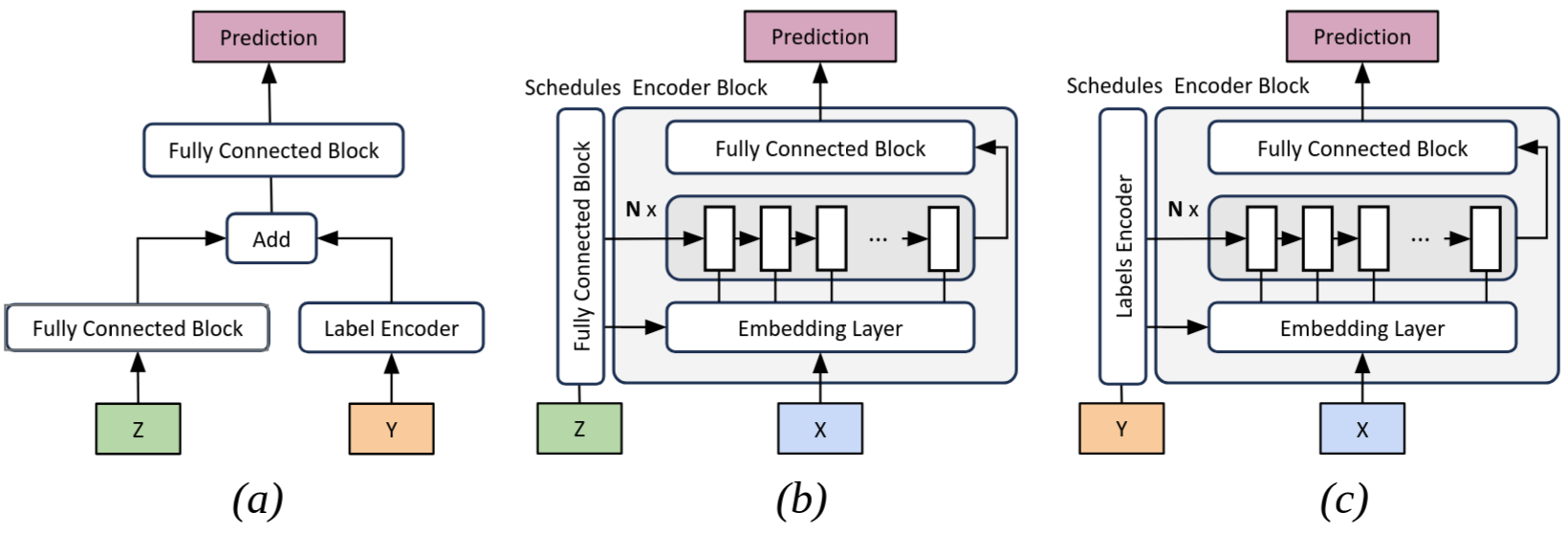}
    \caption{Mutual information estimation (statistics) models: (a) latent-labels $I(z; y)$, (b) latent-schedules $I(z; x)$, and (c) schedules-labels $I(x; y)$}
    \label{fig:statistics_networks}
\end{figure}

We employ statistics models as per Figure \ref{fig:statistics_networks}. All models are based on components used for the ActVAE model, detailed in Section \ref{sec:components}. We use a hidden size of 256 and depth of 2 throughout for all LSTM blocks and feed-forward blocks. Feed forwards blocks are composed of two linear layers with ReLU activations. All models are trained using pairs of samples from their respective distributions. Modelled latents and their partners are inferred from a test set of schedules and labels. All models are trained on 80\% of their respective samples, with 10\% withheld for validation and early stopping. We report all estimates of MI in our results (Section \ref{sec:latent-label_mi_measurement}) using a withheld 10\% test set.

\section{Label-level and Category-level density estimations}
\label{app:cat-level}

This section presents decompositions of the joint density estimation framework into labels (Table \ref{tab:label_fidelity}) and then categories split into the participations (Table \ref{tab:cat_participation_EMD}), transitions (Table \ref{tab:cat_transitions_EMD}) and timing (Table \ref{tab:cat_timing_EMD}) domains.

Domain distances are measured as EMD. This captures the difference between distributions, not limited to expectation shift but can make interpretation difficult. In the case of participations and transitions, the unit of EMD is \emph{rate}. A distance of 0.05 can then be considered as an expectation that an activity or collection or sequence of activities will be 5\% over or under estimated. Note that the transitions domain includes bi-grams, tri-grams, and quad-grams which have higher cardinality than participations. Therefore the weighted average of distances is naturally lower than for participations. The unit of EMD for timing is \emph{days}, either of duration or start-time or in combination. A typical ActVAE timing distance of 0.02 can be considered as an expectation that an activity will be 30 minutes too long or short, or early or late.

\begin{table}
    \footnotesize
    \caption{Label-level joint density estimation}
    \vspace{2ex}
    \centering
    
        \begin{tabular}{l | c | c | c | c }
 \hline
 & Compositional* & ActVAE-Cond* & ActVAE-Gen* & ActVAE* \\
 \hline \hline
 \multicolumn{5}{l}{}\\[-1em]
\multicolumn{5}{l}{\textbf{Participation domain $\downarrow$}} \\
\hline
Age & \textbf{0.047} & 0.377 & 0.109 & 0.052 \\
Day & 0.045 & 0.372 & 0.066 & \textbf{0.043} \\
Employment & \textbf{0.047} & 0.379 & 0.103 & 0.053 \\
HH income & 0.045 & 0.371 & 0.064 & \textbf{0.040} \\
HH zone & 0.045 & 0.370 & 0.068 & \textbf{0.040} \\
PT A/E distance & 0.046 & 0.370 & 0.088 & \textbf{0.045} \\
Sex & 0.045 & 0.370 & 0.060 & \textbf{0.040} \\
Vehicles & 0.349 & 0.609 & 0.382 & \textbf{0.349} \\
\hline \hline
\multicolumn{5}{l}{}\\[-1em]
\multicolumn{5}{l}{\textbf{Transitions domain $\downarrow$}} \\
\hline
Age & \textbf{0.003} & 0.021 & 0.010 & 0.004 \\
Day & \textbf{0.003} & 0.020 & 0.005 & 0.004 \\
Employment & \textbf{0.003} & 0.021 & 0.010 & 0.004 \\
HH income & 0.003 & 0.020 & 0.005 & \textbf{0.003} \\
HH zone & 0.003 & 0.020 & 0.004 & \textbf{0.003} \\
PT A/E distance & 0.003 & 0.020 & 0.005 & \textbf{0.003} \\
Sex & 0.003 & 0.020 & 0.004 & \textbf{0.003} \\
Vehicles & 0.007 & 0.022 & 0.008 & \textbf{0.007} \\
\hline \hline
\multicolumn{5}{l}{}\\[-1em]
\multicolumn{5}{l}{\textbf{Timing domain $\downarrow$}} \\
\hline
Age & 0.056 & 0.192 & 0.047 & \textbf{0.025} \\
Day & 0.055 & 0.172 & 0.027 & \textbf{0.021} \\
Employment & 0.055 & 0.188 & 0.044 & \textbf{0.024} \\
HH income & 0.054 & 0.168 & 0.025 & \textbf{0.019} \\
HH zone & 0.053 & 0.170 & 0.023 & \textbf{0.019} \\
PT A/E distance & 0.054 & 0.174 & 0.027 & \textbf{0.020} \\
Sex & 0.053 & 0.169 & 0.022 & \textbf{0.018} \\
Vehicles & 0.258 & 0.366 & 0.235 & \textbf{0.229} \\

        \hline
        \multicolumn{5}{l}{}\\[-1em]
        \multicolumn{5}{l}{{\small \textasteriskcentered{ Mean EMD from 5 model runs, variance is $<$0.0005 throughout.}}} \\
        \end{tabular}
    \label{tab:label_fidelity}
\end{table}

\begin{table}
    \footnotesize
    \caption{Category-level participations joint density estimation}
    \vspace{2ex}
    \centering
    
        \begin{tabular}{l | c | c | c | c }
 \hline
 & Compositional* & ActVAE-Cond* & ActVAE-Gen* & ActVAE* \\
 
 \hline \hline
 \multicolumn{5}{l}{}\\[-1em]
\multicolumn{5}{l}{\textbf{Age $\downarrow$}} \\
\hline
0-10 & \textbf{0.030} & 0.275 & 0.114 & 0.042 \\
10-21 & \textbf{0.021} & 0.221 & 0.159 & 0.055 \\
21-32 & \textbf{0.034} & 0.310 & 0.065 & 0.050 \\
32-40 & 0.054 & 0.425 & 0.101 & \textbf{0.049} \\
40-48 & 0.074 & 0.479 & 0.170 & \textbf{0.063} \\
48-54 & 0.069 & 0.439 & 0.109 & \textbf{0.052} \\
54-61 & 0.060 & 0.438 & 0.097 & \textbf{0.050} \\
61-68 & 0.059 & 0.430 & 0.101 & \textbf{0.049} \\
68-75 & \textbf{0.043} & 0.421 & 0.094 & 0.059 \\
75+ & \textbf{0.032} & 0.339 & 0.080 & 0.056 \\
 \hline \hline
\multicolumn{5}{l}{\textbf{Day $\downarrow$}} \\
\hline
Monday & 0.048 & 0.359 & 0.059 & \textbf{0.043} \\
Tuesday & 0.051 & 0.373 & 0.064 & \textbf{0.042} \\
Wednesday & 0.050 & 0.389 & 0.071 & \textbf{0.049} \\
Thursday & 0.051 & 0.382 & 0.066 & \textbf{0.047} \\
Friday & \textbf{0.045} & 0.402 & 0.075 & 0.052 \\
Saturday & \textbf{0.038} & 0.387 & 0.067 & 0.038 \\
Sunday & 0.033 & 0.314 & 0.059 & \textbf{0.031} \\
\hline \hline
 \multicolumn{5}{l}{}\\[-1em]
\multicolumn{5}{l}{\textbf{Employment $\downarrow$}} \\
\hline
ft-employed & 0.052 & 0.388 & 0.079 & \textbf{0.049} \\
other & \textbf{0.068} & 0.468 & 0.166 & 0.087 \\
pt-employed & 0.081 & 0.457 & 0.150 & \textbf{0.056} \\
retired & \textbf{0.036} & 0.409 & 0.085 & 0.058 \\
student & \textbf{0.028} & 0.213 & 0.164 & 0.058 \\
unemployed & \textbf{0.079} & 0.459 & 0.125 & 0.111 \\
 \hline \hline
 \multicolumn{5}{l}{}\\[-1em]
\multicolumn{5}{l}{\textbf{HH income $\downarrow$}} \\
\hline
lowest & 0.038 & 0.314 & 0.054 & \textbf{0.033} \\
low & 0.044 & 0.381 & 0.065 & \textbf{0.040} \\
mid & 0.042 & 0.373 & 0.056 & \textbf{0.036} \\
high & 0.055 & 0.405 & 0.079 & \textbf{0.048} \\
highest & \textbf{0.045} & 0.380 & 0.066 & 0.045 \\
 \hline \hline
 \multicolumn{5}{l}{}\\[-1em]
\multicolumn{5}{l}{\textbf{HH zone $\downarrow$}} \\
\hline
rural & 0.053 & 0.401 & 0.070 & \textbf{0.044} \\
suburban & 0.033 & 0.310 & 0.058 & \textbf{0.032} \\
urban & 0.049 & 0.400 & 0.074 & \textbf{0.045} \\
 \hline \hline
 \multicolumn{5}{l}{}\\[-1em]
\multicolumn{5}{l}{\textbf{PT access/egress distance $\downarrow$}} \\
\hline
closest & \textbf{0.025} & 0.267 & 0.090 & 0.036 \\
close & 0.057 & 0.433 & 0.100 & \textbf{0.053} \\
mid & 0.054 & 0.430 & 0.099 & \textbf{0.050} \\
far & 0.055 & 0.416 & 0.082 & \textbf{0.048} \\
furthest & 0.040 & 0.329 & 0.048 & \textbf{0.036} \\
 \hline \hline
 \multicolumn{5}{l}{}\\[-1em]
\multicolumn{5}{l}{\textbf{Sex $\downarrow$}} \\
\hline
female & 0.049 & 0.383 & 0.063 & \textbf{0.041} \\
male & 0.039 & 0.356 & 0.057 & \textbf{0.039} \\
 \hline \hline
 \multicolumn{5}{l}{}\\[-1em]
\multicolumn{5}{l}{\textbf{Vehicles $\downarrow$}} \\
\hline
0 & \textbf{0.671} & 0.764 & 0.730 & 0.675 \\
1 & 0.776 & 0.947 & 0.786 & \textbf{0.774} \\
2 & \textbf{0.799} & 0.981 & 0.819 & 0.799 \\
3 & \textbf{0.789} & 0.961 & 0.800 & 0.791 \\
        \hline
        \multicolumn{5}{l}{}\\[-1em]
        \multicolumn{5}{l}{{\small \textasteriskcentered{ Mean EMD from 5 model runs, variance is $<$0.0005 throughout.}}} \\
        \end{tabular}
    \label{tab:cat_participation_EMD}
\end{table}

\begin{table}
    \footnotesize
    \caption{Category-level transitions joint density estimation}
    \vspace{2ex}
    \centering
    
        \begin{tabular}{l | c | c | c | c }
 \hline
 & Compositional* & ActVAE-Cond* & ActVAE-Gen* & ActVAE* \\
 
 \hline \hline
 \multicolumn{5}{l}{}\\[-1em]
\multicolumn{5}{l}{\textbf{Age $\downarrow$}} \\
\hline
0-10 & \textbf{0.003} & 0.016 & 0.015 & 0.004 \\
10-21 & \textbf{0.002} & 0.013 & 0.013 & 0.004 \\
21-32 & \textbf{0.003} & 0.017 & 0.008 & 0.004 \\
32-40 & 0.004 & 0.022 & 0.008 & \textbf{0.004} \\
40-48 & \textbf{0.004} & 0.023 & 0.009 & 0.004 \\
48-54 & \textbf{0.004} & 0.021 & 0.007 & 0.004 \\
54-61 & 0.004 & 0.022 & 0.007 & \textbf{0.003} \\
61-68 & \textbf{0.004} & 0.024 & 0.009 & 0.004 \\
68-75 & \textbf{0.004} & 0.026 & 0.011 & 0.005 \\
75+ & \textbf{0.003} & 0.022 & 0.013 & 0.005 \\
 \hline \hline
\multicolumn{5}{l}{\textbf{Day $\downarrow$}} \\
\hline
Monday & \textbf{0.003} & 0.019 & 0.005 & 0.003 \\
Tuesday & \textbf{0.003} & 0.019 & 0.005 & 0.003 \\
Wednesday & \textbf{0.003} & 0.019 & 0.005 & 0.003 \\
Thursday & 0.003 & 0.019 & 0.004 & \textbf{0.003} \\
Friday & 0.003 & 0.020 & 0.005 & \textbf{0.003} \\
Saturday & \textbf{0.003} & 0.026 & 0.009 & 0.004 \\
Sunday & \textbf{0.003} & 0.020 & 0.006 & 0.004 \\
\hline \hline
 \multicolumn{5}{l}{}\\[-1em]
\multicolumn{5}{l}{\textbf{Employment $\downarrow$}} \\
\hline
ft-employed & \textbf{0.003} & 0.022 & 0.008 & 0.003 \\
other & \textbf{0.006} & 0.024 & 0.012 & 0.008 \\
pt-employed & 0.004 & 0.020 & 0.007 & \textbf{0.004} \\
retired & \textbf{0.003} & 0.026 & 0.012 & 0.005 \\
student & \textbf{0.003} & 0.011 & 0.014 & 0.005 \\
unemployed & \textbf{0.007} & 0.026 & 0.012 & 0.008 \\
 \hline \hline
 \multicolumn{5}{l}{}\\[-1em]
\multicolumn{5}{l}{\textbf{HH income $\downarrow$}} \\
\hline
lowest & 0.003 & 0.019 & 0.005 & \textbf{0.003} \\
low & 0.003 & 0.021 & 0.005 & \textbf{0.003} \\
mid & 0.003 & 0.021 & 0.004 & \textbf{0.003} \\
high & 0.003 & 0.020 & 0.005 & \textbf{0.003} \\
highest & 0.003 & 0.019 & 0.006 & \textbf{0.003} \\
 \hline \hline
 \multicolumn{5}{l}{}\\[-1em]
\multicolumn{5}{l}{\textbf{HH zone $\downarrow$}} \\
\hline
rural & 0.003 & 0.021 & 0.005 & \textbf{0.003} \\
suburban & \textbf{0.003} & 0.018 & 0.004 & 0.003 \\
urban & 0.003 & 0.021 & 0.004 & \textbf{0.003} \\
 \hline \hline
 \multicolumn{5}{l}{}\\[-1em]
\multicolumn{5}{l}{\textbf{PT access/egress distance $\downarrow$}} \\
\hline
closest & \textbf{0.003} & 0.016 & 0.005 & 0.004 \\
close & 0.004 & 0.022 & 0.005 & \textbf{0.004} \\
mid & 0.003 & 0.021 & 0.005 & \textbf{0.003} \\
far & 0.003 & 0.021 & 0.005 & \textbf{0.003} \\
furthest & 0.003 & 0.019 & 0.005 & \textbf{0.003} \\
 \hline \hline
 \multicolumn{5}{l}{}\\[-1em]
\multicolumn{5}{l}{\textbf{Sex $\downarrow$}} \\
\hline
female & 0.003 & 0.020 & 0.004 & \textbf{0.003} \\
male & 0.003 & 0.019 & 0.005 & \textbf{0.003} \\
 \hline \hline
 \multicolumn{5}{l}{}\\[-1em]
\multicolumn{5}{l}{\textbf{Vehicles $\downarrow$}} \\
\hline
0 & \textbf{0.009} & 0.016 & 0.010 & 0.010 \\
1 & 0.011 & 0.022 & 0.012 & \textbf{0.011} \\
2 & 0.012 & 0.022 & 0.013 & \textbf{0.012} \\
3 & \textbf{0.012} & 0.020 & 0.013 & 0.012 \\
        \hline
        \multicolumn{5}{l}{}\\[-1em]
        \multicolumn{5}{l}{{\small \textasteriskcentered{ Mean EMD from 5 model runs, variance is $<$0.0005 throughout.}}} \\
        \end{tabular}
    \label{tab:cat_transitions_EMD}
\end{table}

\begin{table}
    \footnotesize
    \caption{Category-level timing joint density estimation}
    \vspace{2ex}
    \centering
    
        \begin{tabular}{l | c | c | c | c }
 \hline
 & Compositional* & ActVAE-Cond* & ActVAE-Gen* & ActVAE* \\
 
 \hline \hline
 \multicolumn{5}{l}{}\\[-1em]
\multicolumn{5}{l}{\textbf{Age $\downarrow$}} \\
\hline
0-10 & 0.061 & 0.199 & 0.079 & \textbf{0.025} \\
10-21 & 0.065 & 0.164 & 0.052 & \textbf{0.033} \\
21-32 & 0.063 & 0.182 & 0.042 & \textbf{0.026} \\
32-40 & 0.059 & 0.190 & 0.037 & \textbf{0.024} \\
40-48 & 0.058 & 0.186 & 0.040 & \textbf{0.023} \\
48-54 & 0.058 & 0.191 & 0.035 & \textbf{0.022} \\
54-61 & 0.054 & 0.202 & 0.028 & \textbf{0.021} \\
61-68 & 0.053 & 0.200 & 0.033 & \textbf{0.020} \\
68-75 & 0.048 & 0.206 & 0.060 & \textbf{0.026} \\
75+ & 0.044 & 0.195 & 0.070 & \textbf{0.028} \\
 \hline \hline
\multicolumn{5}{l}{\textbf{Day $\downarrow$}} \\
\hline
Monday & 0.054 & 0.164 & 0.027 & \textbf{0.022} \\
Tuesday & 0.053 & 0.165 & 0.029 & \textbf{0.022} \\
Wednesday & 0.054 & 0.163 & 0.028 & \textbf{0.022} \\
Thursday & 0.055 & 0.167 & 0.027 & \textbf{0.021} \\
Friday & 0.055 & 0.173 & 0.025 & \textbf{0.020} \\
Saturday & 0.059 & 0.198 & 0.027 & \textbf{0.019} \\
Sunday & 0.052 & 0.176 & 0.029 & \textbf{0.021} \\
\hline \hline
 \multicolumn{5}{l}{}\\[-1em]
\multicolumn{5}{l}{\textbf{Employment $\downarrow$}} \\
\hline
ft-employed & 0.059 & 0.188 & 0.042 & \textbf{0.024} \\
other & 0.063 & 0.177 & 0.040 & \textbf{0.032} \\
pt-employed & 0.052 & 0.187 & 0.032 & \textbf{0.018} \\
retired & 0.048 & 0.197 & 0.051 & \textbf{0.024} \\
student & 0.052 & 0.171 & 0.048 & \textbf{0.033} \\
unemployed & 0.062 & 0.215 & 0.045 & \textbf{0.036} \\
 \hline \hline
 \multicolumn{5}{l}{}\\[-1em]
\multicolumn{5}{l}{\textbf{HH income $\downarrow$}} \\
\hline
lowest & 0.049 & 0.154 & 0.025 & \textbf{0.018} \\
low & 0.052 & 0.168 & 0.022 & \textbf{0.019} \\
mid & 0.055 & 0.171 & 0.022 & \textbf{0.019} \\
high & 0.055 & 0.171 & 0.027 & \textbf{0.020} \\
highest & 0.058 & 0.175 & 0.030 & \textbf{0.020} \\
 \hline \hline
 \multicolumn{5}{l}{}\\[-1em]
\multicolumn{5}{l}{\textbf{HH zone $\downarrow$}} \\
\hline
rural & 0.053 & 0.184 & 0.023 & \textbf{0.019} \\
suburban & 0.054 & 0.164 & 0.024 & \textbf{0.019} \\
urban & 0.053 & 0.168 & 0.023 & \textbf{0.018} \\
 \hline \hline
 \multicolumn{5}{l}{}\\[-1em]
\multicolumn{5}{l}{\textbf{PT access/egress distance $\downarrow$}} \\
\hline
closest & 0.049 & 0.152 & 0.030 & \textbf{0.021} \\
close & 0.053 & 0.176 & 0.027 & \textbf{0.019} \\
mid & 0.053 & 0.179 & 0.025 & \textbf{0.018} \\
far & 0.057 & 0.184 & 0.024 & \textbf{0.019} \\
furthest & 0.060 & 0.181 & 0.028 & \textbf{0.021} \\
 \hline \hline
 \multicolumn{5}{l}{}\\[-1em]
\multicolumn{5}{l}{\textbf{Sex $\downarrow$}} \\
\hline
female & 0.054 & 0.166 & 0.021 & \textbf{0.018} \\
male & 0.053 & 0.171 & 0.024 & \textbf{0.018} \\
 \hline \hline
 \multicolumn{5}{l}{}\\[-1em]
\multicolumn{5}{l}{\textbf{Vehicles $\downarrow$}} \\
\hline
0 & 0.526 & 0.569 & 0.517 & \textbf{0.511} \\
1 & 0.527 & 0.582 & 0.510 & \textbf{0.509} \\
2 & 0.527 & 0.583 & 0.514 & \textbf{0.510} \\
3 & 0.531 & 0.595 & 0.518 & \textbf{0.512} \\
        \hline
        \multicolumn{5}{l}{}\\[-1em]
        \multicolumn{5}{l}{{\small \textasteriskcentered{ Mean EMD from 5 model runs, variance is $<$0.0005 throughout.}}} \\
        \end{tabular}
    \label{tab:cat_timing_EMD}
\end{table}

\section*{CRediT authorship contribution statement}

\textbf{Fred Shone:} Conceptualization, Methodology, Software, Formal analysis, Investigation, Writing – original draft, Writing – review \& editing, Visualization. 
\textbf{Tim Hillel:} Conceptualization, Methodology, Writing - review \& editing, Supervision, Funding acquisition. 

\section*{Acknowledgments}

This work was supported by the Engineering and Physical Sciences Research Council (EPSRC) [grant numbers EP/T517793/1 and EP/W524335/1].


\bibliographystyle{elsarticle-harv}
\bibliography{cas-refs}





\end{document}